\documentclass{article}
\ifdefined\pdfminorversion\pdfminorversion=7\fi

\usepackage[]{iclr2027_conference,times}
\iclrfinalcopy 
\usepackage[T1]{fontenc}
\usepackage{amsmath,amssymb,amsthm,mathtools,bm}
\usepackage{graphicx,booktabs,array,multirow}
\usepackage{arydshln}
\usepackage{algorithm,algpseudocode}
\usepackage{xcolor,hyperref,url,microtype}
\usepackage[export]{adjustbox}
\usepackage[]{changes}

\definechangesauthor[name=Yuxuan]{YQ}
\renewcommand{\added}[2][]{#2}
\renewcommand{\deleted}[2][]{}
\renewcommand{\replaced}[3][]{#2}

\usepackage{wrapfig,needspace,float,placeins}
\hypersetup{hidelinks}
\graphicspath{{Figs/}}
\newtheorem{proposition}{Proposition}[section]

\newcommand{\mcl}[1]{\mathcal{#1}}

\newcommand{\mbf}[1]{\mathbf{#1}}
\newcommand{\n}{\mathbf n}
\newcommand{\z}{\mathbf z}
\newcommand{\vv}{\mathbf v}
\newcommand{\w}{\mathbf w}
\newcommand{\Sm}{\mathbb S^{-}}
\newcommand{\Sp}{\mathbb S^{+}}
\newcommand{\St}{\mathbb S}
\newcommand{\mud}{\widehat{\boldsymbol\mu}}

\newcommand{\Dref}{\mcl D_R}
\newcommand{\Dtilt}{\mcl D_\theta^{\,\mathrm{tilt}}}
\newcommand{\Qtilt}{\widehat Q_\theta^{\,\mathrm{tilt}}}
\DeclareMathOperator{\Cov}{Cov}

\DeclareMathOperator{\diag}{diag}
\DeclareMathOperator{\ESS}{ESS}

\newcommand{\topic}[1]{\par\smallskip\noindent\textbf{#1.}\ }

\newcommand{\floattoptrim}{\vspace*{-0.65\baselineskip}}
\title{CRNDiff: Count-Native Diffusion Framework via Chemical Reaction Networks}

\author{
  Yuxuan Qiu,
  Praful Gagrani 
 \&
  Tetsuya J Kobayashi\\
Institute of Industrial Science\\
The University of Tokyo\\
Tokyo 153-8505, Japan\\
  \texttt{qiu@sat.t.u-tokyo.ac.jp}
}
\date{}
\begin{document}
\maketitle
\lhead{}

\begin{abstract}
Scientific measurements such as single-cell RNA (scRNA) sequencing often take the form of
nonnegative integer counts,
\replaced[id=YQ]
{whereas continuous-state diffusion models approximate this discrete structure using continuous coordinates}
{but continuous-state diffusion models may not represent them naturally since they leave the integer lattice}.
Building on stochastic chemical reaction networks (CRNs), a class of count-native
Markov jump processes, we introduce CRNDiff, a structured framework that combines
count-space diffusion with inference-time conditioning on rare subpopulations.
An independent birth--death instantiation yields a closed-form transition
kernel for forward noising. This kernel enables reverse sampling via
forward-filtering backward-sampling (FFBS) and supports data-driven selection
of the terminal noising time, eliminating the need for a validation sweep. This
tractability also lets us introduce tilted Feynman--Kac (FK) steering, a method for
sampling target subpopulations from a frozen generator without retraining. By tilting
posterior marginals before FK particle correction, steering mitigates
importance-weight concentration when the target population is rare.
Using \replaced[id=YQ]{scRNA-seq data from}{the scRNA-seq data of} the human heart cell atlas,
we test the ability of CRNDiff to generate cell-type-specific distributions.
Across the three evaluated target populations, CRNDiff achieves the highest conditional
fidelity among the \replaced[id=YQ]{evaluated generative models}{competing generative models evaluated},
with larger mean purity margins for rarer target populations.
Generated cells preserve marker-level differential-expression structure.
Replacing real training cells for the target classes with generated cells yields downstream
classification performance approaching that of the real-data reference.
\end{abstract}

\section{Introduction}

Diffusion models generate data by reversing a stochastic process that progressively destroys its
structure~\citep{ho2020denoising,song2020score}. The forward process is therefore a modeling choice,
and its design should reflect the state space and the meaningful local transformations of the data.
We focus on nonnegative integer counts. In single-cell RNA sequencing, for example, each coordinate
records a transcript count and lies on the nonnegative integer lattice~\citep{lopez2018deep}.
\replaced[id=YQ]
{While finite-state discrete diffusion models operate on a fixed, finite set of states}
{Finite-state discrete diffusion models operate on a fixed, finite set of states}
~\citep{austin2023structureddenoisingdiffusionmodels,campbell2022continuoustimeframeworkdiscrete},
\replaced[id=YQ]
{nonnegative integer counts additionally have a natural ordering and no intrinsic upper bound.}
{Nonnegative integer counts, however, have a natural ordering and no intrinsic upper bound.}
\replaced[id=YQ]
{To accommodate these properties, recent count-native models formulate diffusion directly on the
nonnegative integer lattice, a countably infinite state space.}
{Recent count-native models therefore formulate diffusion on the nonnegative integer lattice,
a countably infinite state space.}
Blackout Diffusion and CountsDiff employ pure-death or thinning dynamics~
\citep{santos2023blackoutdiffusiongenerativediffusion,soatto2026countsdiffdiffusionmodelnatural},
while Count Bridges and Count-FM use birth--death dynamics~
\citep{fishman2026countbridgesenablemodeling,wei2026flowmatchingcountdata}.
\replaced[id=YQ]
{However, existing count-native diffusion models are formulated around specific transition mechanisms.}
{Existing count-native diffusion models are formulated around specific transition mechanisms.}
A more general, structured framework is needed to specify diverse admissible state transitions and their
state-dependent rates.

Stochastic chemical reaction networks (CRNs) provide such a framework. Reaction channels specify admissible changes in count-valued states, while propensity functions determine their state-dependent rates; stochastic mass-action kinetics further arise naturally from \replaced[id=YQ]{combinatorial interactions among discrete reactant molecules}{combinatorial selection of discrete reactants}~\citep{GILLESPIE1992404}. CRNs thus provide a compositional parameterization of count-space Markov dynamics together with useful analytic structure. Their master equations admit probability-generating-function representations~\citep{doi1976second,peliti1985path,baez2018quantum} that, when tractable, yield exact transition kernels and Markov bridges~\citep{jahnke2007solving,vastola2021solving}. Building on this structure, we introduce CRNDiff and instantiate it with an independent birth--death CRN. The forward process preserves empirical coordinate means and converges to a product-Poisson stationary distribution. Its closed-form transition kernels support bridge-based reverse sampling. The explicit factorial-cumulant dynamics yield a data-driven criterion for selecting the terminal time of the forward process based on excess-covariance decay, without requiring a validation sweep.

This tractable birth--death construction also supports tilted Feynman--Kac steering for inference-time conditioning of a frozen generator~\citep{wu2024practicalasymptoticallyexactconditional,pmlr-v267-singhal25b}. For rare targets where steering based on sequential Monte Carlo (SMC) can suffer from severe importance-weight concentration~\citep{delmoral2004feynman,chopin2020introduction}, marginal tilt first adapts the reverse proposal toward the target marginals, and a Feynman--Kac correction of the residual discrepancy then accounts for the remaining joint mismatch along the reverse chain, without retraining the generator.

We evaluate CRNDiff on a controlled synthetic construction, which separates the effects of marginal tilt and residual correction, and on single-cell RNA sequencing data from an adult human heart cell
atlas~\citep{litvivnukova2020cells}, where one frozen generator serves target populations spanning a wide abundance range. CRNDiff outperforms the evaluated baselines across all conditional-fidelity
metrics. Across the three evaluated target populations, its mean purity margin over the baseline with the highest mean purity increases as training abundance decreases. Generated cells preserve cell-type-specific differential-expression rankings and, when used for downstream cell-type classification, yield performance approaching that of a classifier trained entirely on real cells.


\section{Background}

\subsection{Continuous-time diffusion models}
\label{sec:CT_diff_model}

\replaced[id=YQ]
{Let $\mathcal{D}$ be a distribution on the count space $\mathcal{X}$, which in this work is a nonnegative integer lattice.}
{Let $\mathcal{D}$ be a distribution on a countable state space $\mathcal{X}$,}
\replaced[id=YQ]
{Let $\overrightarrow{q}(x_t \mid x_s)$, $s<t$, denote the forward transition kernel of a continuous-time Markov jump process initialized from $q_0=\mathcal{D}$.}
{and let $q_t$ be the single-time marginal of a continuous-time Markov jump process initialized from $q_0=\mathcal{D}$. We denote its forward transition kernel by $\overrightarrow{q}(x_t \mid x_s)$, $s<t$.}
The corresponding single-time marginal is
\[
q_t(x_t)
=
\sum_{x_0 \in \mathcal{X}}
\overrightarrow{q}(x_t \mid x_0)\,\mathcal{D}(x_0).
\]
\replaced[id=YQ]
{We choose the forward process such that $q_t$ converges to a tractable stationary distribution $\pi$.}
{We choose the process such that $q_t$ converges to a tractable stationary distribution $\pi$.}

\topic{Forward and reverse dynamics}
With transition rates \(w(x\to y)\), \(q_t\) satisfies
\begin{equation}
    \partial_t q_t(x)
    =
    \sum_y
    \left[
        w(y\to x)q_t(y)-w(x\to y)q_t(x)
    \right].
    \label{eq:KFE}
\end{equation}
The corresponding generator is
\(
(Lf)(x)=\sum_yw(x\to y)[f(y)-f(x)]
\),
so that
\(
\partial_t\mathbb E[f(X_t)]=\mathbb E[(Lf)(X_t)].
\)
For reverse time \(s\in[0,T]\), the time-reversed process has rates
\begin{equation}
    \widetilde w_s(y\to x)
    =
    w(x\to y)
    \frac{q_{T-s}(x)}{q_{T-s}(y)}.
    \label{eq:reverse_rates}
\end{equation}
It evolves from \(q_T\) back to \(q_0=\mcl D\); in practice \(q_T\) is
replaced by \(\pi\) once the forward process has sufficiently relaxed.

\topic{Forward-filtering backward-sampling}
\replaced[id=YQ]
{For a tractable forward kernel, define the clean-state posterior
$Q(x_0 \mid x_t,t)$. The corresponding Markov bridge
$B(x_s \mid x_0,x_t)$ is the conditional distribution of the intermediate
state $x_s$ given the endpoints $x_0$ and $x_t$:}
{For a tractable forward kernel, define the clean-state posterior and the Markov bridge,
the conditional distribution of an intermediate state given both endpoints,}
\begin{align}
    Q(x_0\mid x_t,t)
    &=
    \frac{
        \overrightarrow q(x_t\mid x_0)\mcl D(x_0)
    }{
        \sum_{x'}\overrightarrow q(x_t\mid x')\mcl D(x')
    },
    \label{eq:posterior}\\
    B(x_s\mid x_0,x_t)
    &=
    \frac{
        \overrightarrow q(x_t\mid x_s)
        \overrightarrow q(x_s\mid x_0)
    }{
        \overrightarrow q(x_t\mid x_0)
    }.
    \label{eq:bridge}
\end{align}
Marginalizing over \(x_0\) gives
\begin{equation}
    \overleftarrow q(x_s\mid x_t)
    =
    \sum_{x_0}Q(x_0\mid x_t,t)B(x_s\mid x_0,x_t)
    =
    \overrightarrow q(x_t\mid x_s)
    \frac{q_s(x_s)}{q_t(x_t)}.
    \label{eq:reverse_kernel}
\end{equation}
Sampling this kernel backward on any time grid gives exact FFBS when the
posterior and terminal distribution are exact
\citep{carter1994gibbs,chopin2020introduction}. We approximate \(Q\) by
a learned posterior \(Q_\theta\) trained by the cross-entropy loss on
\(x_0\sim\mcl D\) and
\(x_t\sim\overrightarrow q(x_t\mid x_0)\).

\subsubsection{Conditional generation}
\label{sec:conditional_background}

Let \(P\) be the distribution at time \(0\) produced by a reverse sampling process, and \(P^{\mathrm{tar}}\) the desired distribution. Assuming
\(P(x)>0\) whenever \(P^{\mathrm{tar}}(x)>0\), define
\begin{equation}
    H(x)
    :=
    \frac{P^{\mathrm{tar}}(x)}{P(x)}.
    \label{eq:fk_endpoint_weight}
\end{equation}

\topic{Endpoint tilting}
For the posterior associated with prior \(P\), changing the endpoint distribution preserves the forward kernel and bridge and reweights the posterior,
\begin{equation}
    Q^H(x_0\mid x_t,t)
    =
    \frac{
        H(x_0)Q(x_0\mid x_t,t)
    }{
        \sum_{x'}H(x')Q(x'\mid x_t,t)
    }.
    \label{eq:tilt_posterior}
\end{equation}
Thus a learned posterior can be adapted to a new endpoint distribution
without retraining the generator. While exact sampling also requires the target
terminal marginal 
\(\pi^T(x) = \sum_{x_0} \overrightarrow q(x_T=x|x_0)P^{\mathrm{tar}}(x_0)\), initialization from the common stationary distribution $\pi$ is the
usual large-\(T\) approximation.

\topic{Feynman--Kac steering}
When \(P^{\mathrm{tar}}\) concentrates on a region with little probability under \(P\), applying \(H\) only at the
endpoint can produce concentrated importance weights. Feynman--Kac (FK)
steering mitigates this by distributing the same correction along the reverse trajectory
\citep{delmoral2004feynman,
wu2024practicalasymptoticallyexactconditional,
pmlr-v267-singhal25b}.

On \(0=t_0<\cdots<t_K=T\), let \(\mathbb P\) be the reverse proposal distribution over paths \(x_{0:K}=(x_{t_0},\ldots,x_{t_K})\). Choose positive potentials \(\psi_j(x_{t_j})\) with
\(\psi_K\equiv1\), \(\psi_0=H\).
The FK path measure is
\begin{equation}
    \mathbb P_{\mathrm{FK}}(dx_{0:K})
    \propto
    \left(\prod_{j=1}^K G_j\right)
    \mathbb P(dx_{0:K}), \qquad \text{where}\qquad    G_j
    =
    \frac{
        \psi_{j-1}(x_{t_{j-1}})
    }{
        \psi_j(x_{t_j})
    }.
\end{equation}
Since the product telescopes to \(H(x_0)\), its endpoint distribution is
proportional to \(P(x_0)H(x_0)\). The intermediate potentials \(\psi_j\) therefore
affect sampling efficiency but not the desired endpoint correction. When \(\psi\) is not available in closed form, particle methods approximate the continuation potential
\(\psi_j(x_{t_j})=\mathbb E[H(x_0)\mid x_{t_j}]\)
on the fly, while propagating, reweighting, and resampling trajectories.

\subsection{Chemical reaction networks}
\label{sec:crn_background}

\topic{Notation} For any \(S\times R\) matrix \(M\), we use \(M_{\cdot,r}\) to denote its
\(r\)-th column, \(M_{i,\cdot}\) its \(i\)-th row, and \(M_{i,r}\) its
\((i,r)\)-th entry. 
For \(\mbf x,\mbf y\in\mathbb R_{\geq0}^S\), we also use the multi-index notation
\(
    \mbf x^\mbf y := \prod_{i=1}^S x_i^{y_i}.
\)

\textbf{Definition.}
Chemical reaction networks (CRNs) provide a structured class of continuous-time
Markov jump processes on the
\replaced[id=YQ]{count-space}{count space}
$\mathcal{X}=\mathbb{Z}_{\geq 0}^{S}$.
A CRN is specified by a set of species
$\mathcal{S}=\{X_1,\ldots,X_S\}$, whose copy numbers define the state
$n=(n_1,\ldots,n_S)\in\mathcal{X}$, together with $R$ reactions.
Let $\mathbf{S}^{-},\mathbf{S}^{+}\in\mathbb{Z}_{\geq 0}^{S\times R}$
denote the input and output stoichiometric matrices, and define the
stoichiometric matrix $\mathbf{S}=\mathbf{S}^{+}-\mathbf{S}^{-}$.
Thus, $(\mathbf{S}^{-})_{i,r}$ and $(\mathbf{S}^{+})_{i,r}$ are the numbers
of molecules of species $X_i$ consumed and produced by reaction $r$,
respectively, while $\mathbf{S}_{\cdot,r}$ is the corresponding change in
\replaced[id=YQ]{the full count vector $n$}{the full count vector}.
Reaction \(r\) is therefore written as
\begin{equation}
    \sum_{i=1}^S (\Sm)_{i,r}X_i
    \longrightarrow
    \sum_{i=1}^S (\Sp)_{i,r}X_i,
    \qquad
    \n\longrightarrow\n+\St_{\cdot,r}.
    \label{eq:crn_reaction}
\end{equation}

\topic{Master equation and mass action}
Each reaction is assigned a propensity \(j_r(n)\), interpreted as its instantaneous firing rate in state \(n\). Starting from a count vector, the system waits a random time, fires one of the available reactions, and updates the state by its stoichiometric change. Reactions whose required inputs are unavailable have zero propensity, so the dynamics remain in
the nonnegative integer orthant.
The CRN therefore defines transition rates
\(
    w(\n\to\n+\Delta)
    =
    \sum_{r:\,\St_{\cdot,r}=\Delta}
    j_r(\n).
\)
Substituting these rates into Eq.~\eqref{eq:KFE} gives the chemical master
equation
\begin{equation}
    \partial_t q_t(\n)
    =
    \sum_{r=1}^R
    \left[
        j_r(\n-\St_{\cdot,r})
        q_t(\n-\St_{\cdot,r})
        -
        j_r(\n)q_t(\n)
    \right],
    \label{eq:CME}
\end{equation}
where terms corresponding to states outside
\(\mathbb Z_{\geq0}^S\) are understood to vanish.

For stochastic mass-action kinetics, the propensities are
\begin{equation}
    j_r(\n)
    =
    k_rV^{1-|\Sm_{\cdot,r}|}
    \prod_i\frac{n_i!}{(n_i-\Sm_{i,r})!}
    \label{eq:mass_action_propensity}
\end{equation} 
with rate constant \(k_r\), system volume \(V\), and reaction order \( |\Sm_{\cdot,r}|=\sum_i\Sm_{i,r}\). Propensities vanish if any reactant count is insufficient. We set \(V=1\); Appendix~\ref{sec:diffusion_connection} gives the large-\(V\) diffusion approximation.

\topic{Generating functions}
Define the probability generating function (PGF)
\(
    g(\z,t)
    =
    \mathbb E[\z^{\n(t)}]
    =
    \sum_{\n\geq0}\z^\n q_t(\n).
\)
Multiplication by \(z_i\) and differentiation by \(z_i\) implement the
addition and removal of count units, giving the standard
creation--annihilation representation
\citep{doi1976second,peliti1985path,baez2018quantum}. For mass-action
kinetics,
\begin{equation}
    \partial_tg(\z,t)
    =
    \sum_r
    k_rV^{1-|\Sm_{\cdot,r}|}
    \left(
        \z^{\Sp_{\cdot,r}}-\z^{\Sm_{\cdot,r}}
    \right)
    \partial_\z^{\Sm_{\cdot,r}}g(\z,t).
    \label{eq:pgf-pde_CRN}
\end{equation}
Here \(\partial_\z^{\mbf a}=\prod_i\partial_{z_i}^{a_i}\); see Appendix~\ref{app:pgf} for the derivation. With
\(g(\z,0)=\z^{\n_0}\), coefficient extraction gives
\(
q_t(\n\mid\n_0)=[\z^\n]g(\z,t)
\),
which supplies the transition kernel required by FFBS.

For \(|\Sm_{\cdot,r}|\leq1\), the PGF equation is first-order and can be treated
by characteristics. This class includes births, deaths, conversions, and
branching reactions, and is considerably richer than the independent
example used below
\citep{jahnke2007solving,vastola2021solving}.

\section{CRN diffusion model}
\label{sec:model}

We instantiate the CRN framework with a simple birth--death process. We use these reactions purely as a noising mechanism, and they provide a natural way to progressively
randomize count-valued data while remaining on the nonnegative integer lattice.
Importantly, the resulting stochastic process is exactly solvable and
gives closed-form transition probabilities. We also explicitly track how statistical structure inherited from the data---such as correlations
 and deviations from independent Poisson   statistics---is
gradually erased by the forward process, and use this decay to determine how
long the data should be noised before reverse sampling begins. Finally, we take the endpoint distribution induced by the CRNDiff generator
and show how it can be tilted toward a prescribed target distribution, enabling
inference-time conditional generation without retraining the underlying model.

\subsection{Noising with a birth--death model}
\label{sec:birth_death}

\topic{Birth--death network}
For each coordinate \(i\in[S]\), consider the ``birth'' and ``death'' reactions that
increase and decrease the count by one unit, respectively:
\begin{equation}
    \varnothing\xrightarrow{\,Vk_i^+\,}X_i,
    \qquad
    X_i\xrightarrow{\,k_i^-n_i\,}\varnothing.
    \label{eq:birth_death_network}
\end{equation}
We set \(V=1\), define
\(
\mu_i=k_i^+/k_i^-
\)
and
\(
\beta_{t,i}=e^{-k_i^-t}
\).
Solving the PGF equation gives the exact transition kernel
\begin{equation}
\begin{aligned}
\overrightarrow q(\n_t\mid\n_0)
&=
\prod_{i=1}^S
\left[
\operatorname{Bin}(n_{0,i},\beta_{t,i})
*
\operatorname{Pois}\!\left(
\mu_i(1-\beta_{t,i})
\right)
\right](n_{t,i})
\\
\added[id=YQ]{
&=
\prod_{i=1}^S
\sum_{m=0}^{n_{t,i}}
\operatorname{Bin}(n_{0,i},\beta_{t,i})(m)\,
\operatorname{Pois}\!\left(
\mu_i(1-\beta_{t,i})
\right)(n_{t,i}-m)
}
\end{aligned}
\label{eq:BD_kernel}
\end{equation}
where \(\operatorname{Bin}(N,p)\) denotes the binomial distribution
\(
\operatorname{Bin}(N,p)(m)
=
\binom{N}{m}p^m(1-p)^{N-m},
\)
and \(\operatorname{Pois}(\lambda)\) denotes the Poisson distribution
\(
\operatorname{Pois}(\lambda)(m)
=
e^{-\lambda}\lambda^m/m!.
\)

Thus, conditional on \(n_{0,i}\), the count \(n_{t,i}\) is distributed as
the sum of a binomial number of surviving initial count units and an
independent Poisson number of newly created count units.
Reverse sampling uses this analytic kernel as the bridge: it samples a
clean-state proposal from the learned posterior, then an intermediate count
state conditioned on that proposal and the current noisy state.
As \(t\to\infty\),
\(
q_t
\to
\pi(\n)=
\prod_i
\deleted[id=YQ]{\operatorname{Pois}(n_i;\mu_i)}
\,
\added[id=YQ]{\operatorname{Pois}(\mu_i)(n_i)}
\).
The PGF derivation is given in
Appendix~\ref{sec:solve_PGF_PDE}.

\topic{Factorial cumulants}
Let \(\Phi_t=\log g_t\). The \(m\)-th factorial cumulant tensor is defined as
\(
    \kappa^{(m)}_{t,i_1\ldots i_m}
    :=
    \left.
    \frac{\partial^m \Phi_t(\z)}
    {\partial z_{i_1}\cdots \partial z_{i_m}}
    \right|_{\z=\bm 1}.
\)
We denote the second factorial cumulant, also called the excess covariance, as
\(
\Psi_{t,ij}:=\kappa^{(2)}_{t,ij}
    =\Cov(n_i(t),n_j(t))
    -
    \delta_{ij}\mathbb E[n_i(t)].
\)
The exact PGF solution then gives
\begin{equation}
    \kappa^{(1)}_{t,i}= \mathbb E[n_i(t)]
    =
    \mu_i(1-\beta_{t,i})
    +
    \beta_{t,i}\mathbb E[n_i(0)],
    \qquad
    \Psi_{t,ij}
    =
    \beta_{t,i}\beta_{t,j}\Psi_{0,ij}.
    \label{eq:birth_death_cumulant}
\end{equation}
More generally, an order-\(m\) factorial cumulant is multiplied by the
corresponding \(m\) survival factors, \(\kappa^{(m)}_{t,i_1\ldots i_m}=\beta_{t,i_1}\ldots\beta_{t,i_m}\kappa^{(m)}_{0,i_1\ldots i_m}\). We choose
\(
k_i^-=1
\)
and
\(
\mu_i=k_i^+=\mathbb E_{\mcl D}[n_i]
\).
The means are then preserved, while every factorial cumulant of order
\(m\geq2\) decays as \(e^{-mt}\).
\topic{Terminal noising time}
\label{sec:terminal-time}
Let
\(
\mud_0=\diag(\mu_1,\ldots,\mu_S)
\)
and define the normalized excess-covariance signal along a direction
\(\vv\in\mathbb R^S\) by
\begin{equation}
    \Sigma_t(\vv)
    =
    \frac{\vv^\top\Psi_t\vv}
         {\vv^\top\mud_0\vv}.
    \label{eq:directional_signal}
\end{equation}
Let \(\w_k\) be a generalized eigenvector of the pair
\((\Psi_0,\mud_0)\), with eigenvalue \(\sigma_k\), so that
\(
    \Psi_0\w_k=\sigma_k\mud_0\w_k.
\)
Since \(\Psi_t=e^{-2t}\Psi_0\), the signal along each generalized
eigendirection decays as
\begin{equation}
    \Sigma_t(\w_k)=e^{-2t}\sigma_k.
    \label{eq:signal_time}
\end{equation}

For a product-Poisson distribution, each coordinate has equal mean and
variance and distinct coordinates are independent, so the population
excess-covariance matrix vanishes, \(\Psi=0\). A finite sample, however,
produces a nonzero apparent excess covariance through sampling
fluctuations alone. We define \(\sigma_{\mathrm{noise}}\) as the spectral
level expected from these finite-sample fluctuations under the
product-Poisson null. This threshold is determined in closed form by the
number of species (genes) and observations (cells), without requiring
Monte Carlo sampling; see Eq.~\eqref{eq:mp_edge}.

Let \(\sigma_1\) denote the largest generalized eigenvalue of
\((\Psi_0,\mud_0)\). We choose the terminal noising time as the time at
which this strongest positive excess-covariance mode reaches the
finite-sample Poisson noise floor,
\begin{equation}
    T_O
    =
    \frac12
    \log\frac{\sigma_1}{\sigma_{\mathrm{noise}}},
    \qquad
    \sigma_1>\sigma_{\mathrm{noise}},
    \label{eq:optimal_terminal_time}
\end{equation}
and set \(T_O=0\) otherwise. Thus, \(T_O\) is the earliest time at which
the strongest positive second-order signal is no longer distinguishable
from finite-sample fluctuations of the product-Poisson null. This is a
second-order calibration criterion, not a guarantee of full distributional
convergence. The spectral
calculation and qualifications are given in
Appendix~\ref{app:signal}.

\subsection{Conditional denoising with tilted FK steering}
\label{sec:fk}

Let $\mathcal D_\theta$ be the endpoint distribution induced by the frozen CRNDiff generator and its unconditional
reverse sampler, $\mathcal D_\theta^{\mathrm{tilt}}$ \added[id=YQ]{be} its marginally tilted counterpart,
and $\mathcal D_R$ \added[id=YQ]{be} the target distribution.
\replaced[id=YQ]
{The two conditioning stages are}
{The two stages are}
\[
\mathcal D_\theta
\xrightarrow{\text{marginal tilt}}
\mathcal D_\theta^{\mathrm{tilt}}
\xrightarrow{\text{residual correction}}
\mathcal D_R.
\]
The first stage uses the factorized posterior to adapt the proposal toward the target marginals;
the second stage corrects the remaining joint discrepancy.
\replaced[id=YQ]
{We call this conditioning procedure tilted FK steering.
Algorithm~\ref{alg:tilted_fk} summarizes the conditioning procedure, while
Algorithm~\ref{alg:train} describes generator training.}
{We call this two-stage sampler tilted FK steering
(Algorithm~2; Algorithm~1 trains the generator).}
The generator parameters remain frozen during conditioning.
Target-specific preparation estimates $\mathcal D_{\theta,i}$ from an unconditional generation
run, estimates $\mathcal D_{R,i}$ from target training cells, and fits the residual discriminator using target cells
and marginally tilted proposal samples.
The final conditional run starts from the terminal distribution
and interleaves tilted reverse transitions with FK weight updates and resampling.
\topic{Marginal tilting}
For high-dimensional count data we sample the learned posterior
coordinate-wise,
\begin{equation}
    \widehat Q_\theta(\n_0\mid\n_t,t)
    =
    \prod_{i=1}^S
    Q_{\theta,i}(n_{0,i}\mid\n_t,t).
    \label{eq:factorized_posterior}
\end{equation}

Each posterior factor
\replaced[id=YQ]{is conditioned on}{conditions on}
the full noisy count vector;
coordinate-wise posterior sampling therefore does not imply a factorized endpoint distribution.

We define the marginally tilted posterior
\begin{equation}
    \Qtilt(\n_0\mid\n_t,t)
    =
    \prod_i
    \frac{
        u_i(n_{0,i})\,
        Q_{\theta,i}(n_{0,i}\mid\n_t,t)
    }{
        \sum_m
        u_i(m)\,
        Q_{\theta,i}(m\mid\n_t,t)
    },
    \label{eq:factorized_weight}
\end{equation}
where the coordinate-wise tilt is
\begin{equation}
    u_i(n_i)
    :=
    \left(
        \frac{\mcl D_{R,i}(n_i)}
             {\mcl D_{\theta,i}(n_i)}
    \right)^{\tau},
    \qquad
    \tau\in(0,1].
    \label{eq:marginal_ratio}
\end{equation}
Here \(\mcl D_{R,i}\) and \(\mcl D_{\theta,i}\) are the \(i\)-th coordinate
marginals of the target and generator endpoint distributions,
respectively. 
Because the endpoint weight
\(\prod_i u_i(n_{0,i})\) factorizes over coordinates, inserting it into
Eq.~\eqref{eq:tilt_posterior} preserves the factorized posterior structure and
tilts each coordinate separately. Combined with the original bridge, this
defines a marginally tilted reverse process with endpoint distribution
\(\Dtilt\). 

Observe that \(u_i\) upweights count values that are more common under the
target distribution and downweights those that are more common under the
original sampler. 
The exponent \(\tau\) controls the strength of the tilt:
\(\tau=1\) applies the full marginal density ratio, while smaller values
temper the correction toward the original posterior. Since these ratios are
less reliably estimated for rare targets, we use stronger
tempering for rarer target populations; see Appendix~\ref{app:tilt-tempering}.
Importantly, tempering changes only the proposal \(\Dtilt\), not the target
distribution \(\Dref\). 

\begin{proposition}[Marginal information limitation]
\label{prop:wall}
 For a fixed $\tau$, targets with identical one-dimensional marginals induce the same
marginally tilted reverse process.
\end{proposition}

The statement follows immediately because such targets give identical
\(u_i\). The tilt therefore adapts the proposal using marginal
information, but cannot by itself distinguish targets that differ only in
their joint dependence. Thus, the proposition concerns the information used
by the marginal tilt ratios, not factorization of the endpoint law.

\topic{Residual correction}
We correct the remaining discrepancy using the FK construction of
Section~\ref{sec:conditional_background}. The discriminator \(r_\phi(\n)\) estimates the probability of the target class \(\Dref\) versus the proposal class \(\Dtilt\).
With equal class priors, the population-optimal discriminator under binary cross-entropy has odds~\citep{goodfellow2014generative}
\begin{equation}
    \rho^*(\n)
    =
    \frac{r^*(\n)}{1-r^*(\n)}
    =
    \frac{\Dref(\n)}
         {\Dtilt(\n)}.
    \label{eq:rho}
\end{equation}
We therefore use
\(
\rho_\phi(\n)=r_\phi(\n)/(1-r_\phi(\n))
\)
as the FK endpoint weight. Applying the potentials of
Section~\ref{sec:conditional_background} to the marginally tilted reverse
chain gives
\begin{equation}
    \mcl D_{\mathrm{FK}}(\n)
    \propto
    \Dtilt(\n)\rho_\phi(\n),
    \label{eq:fk_endpoint_model}
\end{equation}
which equals \(\Dref\) for the exact density ratio.

In practice, we estimate the intermediate FK potentials from predicted
clean endpoints and resample when the effective sample size (ESS) becomes small;
the estimator and particle construction are given in
Appendix~\ref{app:conditional_details}.

\newsavebox{\algboxTrain}\newsavebox{\algboxSample}\newlength{\algboxHt}
\sbox{\algboxTrain}{\begin{minipage}{0.47\linewidth}\footnotesize
\begin{algorithmic}[1]
\Require count data \(\mcl D\) with \(N\) cells and \(S\) genes; posterior network \(Q_\theta\)
\State \(\mu_i\gets\mathbb E_{\mcl D}[n_i]\); \(k_i^-\gets1\), \(k_i^+\gets\mu_i\) \hfill Eq.~\eqref{eq:birth_death_network}
\State \(\sigma_1\gets\max\{\sigma:\Psi_0\w=\sigma\mud_0\w\}\) \hfill Eq.~\eqref{eq:signal_time}
\State \(\sigma_{\mathrm{noise}}\gets(1+\sqrt{S/N})^2-1\) \hfill Eq.~\eqref{eq:mp_edge}
\State \(T_O\gets\max\{\tfrac12\log(\sigma_1/\sigma_{\mathrm{noise}}),\,0\}\) \hfill Eq.~\eqref{eq:optimal_terminal_time}
\While{not converged}
  \State \(\n_0\sim\mcl D\), \(t\sim\mathcal U(0,T_O]\)
  \State \(\n_t\sim\overrightarrow q_t(\cdot\mid\n_0)\) \hfill Eq.~\eqref{eq:BD_kernel}
  \State update \(\theta\) on \(-\sum_i\log Q_{\theta,i}(n_{0,i}\mid\n_t,t)\)
\EndWhile
\State \Return \(Q_\theta\), \(T_O\)
\end{algorithmic}\end{minipage}}
\sbox{\algboxSample}{\begin{minipage}{0.51\linewidth}\footnotesize
\begin{algorithmic}[1]
\Require \(Q_\theta\), \(T_O\); target marginals \(\mcl D_{R,i}\), exponent \(\tau\); odds \(\rho_\phi\) of Eq.~\eqref{eq:rho}; \(M\) particles; resampling steps \(\mathcal C\), ESS threshold \(\vartheta\)
\State \(\mcl D_{\theta,i}\gets\) marginals of an unconditional run of \(Q_\theta\)
\State \(\Qtilt\gets\) tilt \(Q_\theta\) with \(u_i=(\mcl D_{R,i}/\mcl D_{\theta,i})^{\tau}\) \hfill Eq.~\eqref{eq:factorized_weight}
\State \(\n^{(m)}_{t_K}\sim\pi\), \(w^{(m)}\gets1\) for \(m=1,\dots,M\)
\For{\(j=K,\dots,1\)}
  \State \(\n^{(m)}_{t_{j-1}}\sim\) tilted reverse step, all \(m\) \hfill Eqs.~\eqref{eq:bridge},\eqref{eq:factorized_weight}
  \State \(w^{(m)}\gets w^{(m)}\,G_j^{(m)}\), \(G_j=\widehat\psi_{j-1}/\widehat\psi_{j}\) \hfill Eq.~\eqref{eq:fk_increment}
  \State \(\textbf{if }j{-}1\in\mathcal C,\ \ESS<\vartheta M:\ \mathrm{resample}_{w};\ w\gets1\)
\EndFor
\State \Return \(n_{\mathrm{out}}\) of the \(\n^{(m)}_0\), drawn \(\propto w^{(m)}\)
\end{algorithmic}\end{minipage}}
\setlength{\algboxHt}{\dimexpr\ht\algboxSample+\dp\algboxSample\relax}
\ifdim\dimexpr\ht\algboxTrain+\dp\algboxTrain\relax>\algboxHt
  \setlength{\algboxHt}{\dimexpr\ht\algboxTrain+\dp\algboxTrain\relax}\fi

\begin{figure}[!t]
\floattoptrim
\begin{minipage}[t]{0.47\linewidth}
\begin{algorithm}[H]
\caption{CRNDiff training}
\label{alg:train}
\begin{minipage}[t][\algboxHt][t]{\linewidth}\usebox{\algboxTrain}\end{minipage}
\end{algorithm}
\end{minipage}\hfill
\begin{minipage}[t]{0.51\linewidth}
\begin{algorithm}[H]
\caption{Tilted FK steering}
\label{alg:tilted_fk}
\begin{minipage}[t][\algboxHt][t]{\linewidth}\usebox{\algboxSample}\end{minipage}
\end{algorithm}
\end{minipage}
\end{figure}

\section{Related work}
\label{sec:related}

\topic{Discrete and count-native generative models}
Discrete diffusion and its continuous-time formulations generate through Markov transitions on discrete states \citep{austin2023structureddenoisingdiffusionmodels,campbell2022continuoustimeframeworkdiscrete}; masked diffusion language models instantiate absorbing-state transitions \citep{NEURIPS2024_eb0b13cc,shi2025simplifiedgeneralizedmaskeddiffusion}, and discrete flow matching and generator matching separate the probability path from the parameterization of its dynamics \citep{gat2024discreteflowmatching,holderrieth2025generatormatchinggenerativemodeling}. On integer counts, Blackout Diffusion and CountsDiff use death or thinning dynamics \citep{santos2023blackoutdiffusiongenerativediffusion,soatto2026countsdiffdiffusionmodelnatural}, JUMP connects generation to thinning and thickening \citep{pmlr-v202-chen23ap}, and Count Bridges and count-FM employ birth--death jump processes \citep{fishman2026countbridgesenablemodeling,wei2026flowmatchingcountdata}. CRNDiff similarly instantiates a birth--death process as a tractable count-space noising mechanism. More generally, the CRN formulation treats this construction as one instance within a broader class of reaction-network-based forward processes.

\topic{Stochastic kinetics and generating functions}
The operator representation used here originates in Doi--Peliti theory \citep{doi1976second,peliti1985path}; related work connects CRNs, stochastic Petri nets, creation--annihilation operators, and factorial-moment hierarchies \citep{baez2018quantum,smith2017flows}. Our kernels, bridges, and factorial-cumulant dynamics instantiate established exact solutions for monomolecular and first-order systems \citep{jahnke2007solving,vastola2021solving}.

\topic{Conditional generation and inference-time steering}
Best-of-\(N\) selection ranks generated candidates by target score \citep{pmlr-v267-singhal25b}, while value-guided resampling injects target information during reverse sampling \citep{li2024derivativefreeguidancecontinuousdiscrete}. Twisted diffusion sampling and FK steering use sequential Monte Carlo (SMC) potentials to steer a frozen generator \citep{wu2024practicalasymptoticallyexactconditional,pmlr-v267-singhal25b}, but rare targets can cause concentrated importance weights and genealogical degeneracy \citep{delmoral2004feynman,chopin2020introduction,jacob2015path}. CRNDiff adapts the reverse proposal toward target marginals before applying the joint density-ratio correction.

\topic{Single-cell generative models}
scVI, scANVI, and CFGen model single-cell counts through latent representations and count-valued decoders~\citep{lopez2018deep,xu2021probabilistic,palma2025multimodalmultiattributegenerationsingle}; CRNDiff instead models count-valued dynamics directly and conditions target populations at sampling time.


\providecommand{\rev}[1]{#1}
\providecommand{\upd}[1]{#1}
\providecommand{\chg}[1]{#1}

\section{Experiments}
\label{sec:results}

\rev{We evaluate CRNDiff on a synthetic count mixture and human-heart scRNA-seq data, assessing unconditional and conditional generation, genealogical diversity, and downstream cell-type classification. Appendices~\ref{app:toy} and~\ref{app:implementation} provide experimental settings and metric definitions.}

\begin{figure}[!t]
\floattoptrim
\begin{minipage}[t]{0.6\linewidth}
\vspace{0pt}
\centering
\includegraphics[width=\linewidth]{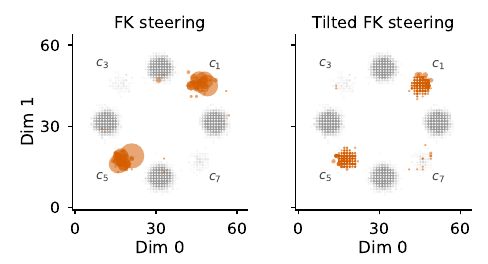}
\end{minipage}\hfill
\begin{minipage}[t]{0.36\linewidth}
\vspace{0pt}
\caption{Generated samples from FK steering and tilted FK steering targeting \(c_1,c_5\). The alternative pair \(c_3,c_7\) has identical one-dimensional marginals. Gray markers show training samples; orange markers show one generated representative per initial ancestor, sized by output multiplicity.}
\label{fig:toy-delivered}
\end{minipage}
\vspace{-0.5em}
\end{figure}

\subsection{A count mixture with matched-marginal targets}
\label{sec:toy}

\rev{We construct an eight-component count mixture with four abundant and four rare components (Figure~\ref{fig:toy-delivered}). The target assigns equal mass to rare components \(c_1\) and \(c_5\), together \(1\%\) of the training mixture; the alternative pair \(c_3,c_7\) has identical one-dimensional marginals but different joint structure.}

\begin{wraptable}{r}{0.42\textwidth}
\vspace{-0.8\baselineskip}
\centering\footnotesize
\setlength{\abovecaptionskip}{0pt}
\setlength{\belowcaptionskip}{3pt}

\caption{Comparison on the two-dimensional count mixture; metrics are defined in Appendix~\ref{app:toy}.}
\label{tab:toy-quality}

\begin{adjustbox}{max width=\linewidth}
\begin{tabular}{lccc}
\toprule
Method & ED $\downarrow$ & mTV $\downarrow$ & Anc. $\uparrow$ \\
\midrule
FK steering & $0.255$ & $0.352$ & $0.116$ \\
Tilted FK steering & $\mathbf{0.036}$ & $\mathbf{0.117}$ & $\mathbf{0.875}$ \\
\midrule
i.i.d.\ target & $0.017$ & $0.050$ & $1.000$ \\
\bottomrule
\end{tabular}
\end{adjustbox}

\end{wraptable}
\vspace{-0.6\baselineskip}

\noindent
On the same frozen generator, FK steering
\citep{wu2024practicalasymptoticallyexactconditional,pmlr-v267-singhal25b}
and tilted FK steering use untilted and marginally tilted proposals, respectively.
Table~\ref{tab:toy-quality} reports energy distance (ED), mean marginal total variation (mTV), and genealogical diversity (Anc.). Here Anc.\ is the number of distinct initial ancestors in the output divided by output sample size. The independent and identically distributed (i.i.d.) row is the target reference. Tilted FK steering reduces the genealogical concentration under FK steering and approaches the reference on all three measures. Appendix~\ref{app:toy} shows that marginal tilt alone cannot distinguish the matched-marginal component pairs and illustrates how marginal tilt and FK steering alter the reverse-chain particle distribution.
\par

\FloatBarrier

\subsection{Generation and evaluation on single-cell data}
\label{sec:hvg}\label{sec:results-singlecell}

We compare CRNDiff with scVI, scANVI, CFGen, and MDLM on
\replaced[id=YQ]
{scRNA-seq data from endothelial, myeloid, and neuronal cells in the adult human heart atlas}
{endothelial, myeloid, and neuronal cells from the adult human heart atlas}
\citep{litvivnukova2020cells,lopez2018deep,xu2021probabilistic,palma2025multimodalmultiattributegenerationsingle,NEURIPS2024_eb0b13cc}.
The CRNDiff generator is trained unconditionally; tilted FK steering conditions its samples on the target cell type.
All fidelity evaluations use the held-out test set as the real-data reference.
Appendix~\ref{app:implementation} details preprocessing, model configurations, sampling, and evaluation.


\paragraph{Conditional fidelity and steering.}

Table~\ref{tab:hvg_single_type_main} evaluates conditional fidelity using purity, mean per-gene Wasserstein-1 distance (\(W_1\)), squared maximum mean discrepancy (\(\mathrm{MMD}^2\)), and Pearson correlation coefficient (PCC).
Purity is the fraction of generated cells assigned to the target population by a fixed external CellTypist heart model~\citep{dominguezconde2022cross}.
\(W_1\), \(\mathrm{MMD}^2\), and PCC compare generated and test cells at the per-gene, joint-distribution, and mean-expression levels, respectively.
CRNDiff has the highest mean purity and PCC, and the lowest mean \(W_1\) and \(\mathrm{MMD}^2\), for all three targets.
Its mean purity margin over the strongest baseline increases as training abundance decreases, and neuronal mean purity approaches the real-cell reference.

\replaced[id=YQ]
{To isolate conditioning effects from generator quality,}
{Because Table~\ref{tab:hvg_single_type_main} compares complete generative methods,}
Tables~\ref{tab:hvg_ablation},~\ref{tab:hvg_tau}, and~\ref{tab:hvg_sampler_axis}
\replaced[id=YQ]
{evaluate conditioning choices}
{isolate conditioning choices}
on the frozen CRNDiff generator
(Appendices~\ref{app:hvg-sampler} and~\ref{app:hvg-sampler-diagnostics}).
Larger best-of-\(N\) pools and stronger value-guided resampling can increase purity and sometimes match or exceed tilted FK steering, often with higher \(W_1\) and \(\mathrm{MMD}^2\).
Ablations show that tilted FK steering preserves more distinct initial ancestors across targets, with metric-dependent distributional trade-offs.

\paragraph{Unconditional generation.}
\replaced[id=YQ]{While}{Because}
Table~\ref{tab:hvg_single_type_main} reflects both generator quality and conditioning,
\replaced[id=YQ]
{Table~\ref{tab:hvg_uncond_fidelity}, which evaluates generators without conditional steering, reports}
{Table~\ref{tab:hvg_uncond_fidelity} evaluates generators without conditional steering. It reports}
sliced Wasserstein-1 distance (sliced-\(W_1\)), median Fano factor, coefficients of variation for library size and detected-gene counts, and CellTypist-assigned target fractions.
CRNDiff attains the lowest sliced-\(W_1\) and near-reference detected-gene variability, whereas lower library-size variability and median Fano factor indicate generator-side dispersion limits.

\paragraph{Marker preservation and downstream utility.}
\added[id=YQ]{In addition,}
Table~\ref{tab:hvg_de} compares differential-expression rankings from generated and real test cells;
CRNDiff has the highest mean top-100 marker overlap and Spearman correlation for endothelial and myeloid cells, while neuronal results are metric-dependent.
Table~\ref{tab:hvg_tstr} evaluates generated cells as replacement training data for the three target classes in an eleven-class classifier.
Class-specific \(F_1\) is the harmonic mean of precision and recall, and macro \(F_1\) is the unweighted mean across all eleven classes.
CRNDiff achieves the highest mean macro \(F_1\), myeloid \(F_1\), and neuronal \(F_1\).


\begin{table}[!t]
\floattoptrim
\centering
\setlength{\abovecaptionskip}{4pt}
\setlength{\belowcaptionskip}{3pt}
\caption{Conditional fidelity for endothelial, myeloid, and neuronal targets. Distributional metrics use held-out target cells from the test set as the real-data reference; purity uses fixed CellTypist target-label mappings. Boldface indicates the best generative-model mean.}
\label{tab:hvg_single_type_main}
\renewcommand{\arraystretch}{0.94}
\begin{tabular}{lcccc}
\toprule
Method
& Purity $\uparrow$
& $W_1$ $\downarrow$
& MMD$^2$ $\downarrow$
& PCC $\uparrow$ \\
\midrule
\multicolumn{5}{l}{
\textbf{Endothelial}\quad
\emph{abundant}, $80{,}463$ training cells
} \\
Real cells
& $0.944$
& $0.011$
& $0.0001$
& $1.000$ \\
\textbf{CRNDiff}
& $\mathbf{0.906}_{\pm.005}$
& $\mathbf{0.064}_{\pm.003}$
& $\mathbf{0.0011}_{\pm.0001}$
& $\mathbf{0.998}_{\pm.000}$ \\
MDLM~\citep{NEURIPS2024_eb0b13cc}
& $0.898_{\pm.026}$
& $0.078_{\pm.029}$
& $0.0058_{\pm.0010}$
& $0.991_{\pm.003}$ \\
CFGen~\citep{palma2025multimodalmultiattributegenerationsingle}
& $0.797_{\pm.009}$
& $0.102_{\pm.001}$
& $0.0036_{\pm.0000}$
& $0.993_{\pm.000}$ \\
scANVI~\citep{xu2021probabilistic}
& $0.857_{\pm.016}$
& $0.079_{\pm.007}$
& $0.0016_{\pm.0001}$
& $0.997_{\pm.000}$ \\
scVI~\citep{lopez2018deep}
& $0.423_{\pm.014}$
& $0.241_{\pm.004}$
& $0.0516_{\pm.0002}$
& $0.862_{\pm.001}$ \\
\midrule
\multicolumn{5}{l}{
\textbf{Myeloid}\quad
$18{,}422$ training cells
} \\
Real cells
& $0.929$
& $0.037$
& $0.0003$
& $0.999$ \\
\textbf{CRNDiff}
& $\mathbf{0.886}_{\pm.014}$
& $\mathbf{0.110}_{\pm.017}$
& $\mathbf{0.0032}_{\pm.0010}$
& $\mathbf{0.987}_{\pm.005}$ \\
MDLM
& $0.840_{\pm.072}$
& $0.147_{\pm.012}$
& $0.0060_{\pm.0027}$
& $0.979_{\pm.014}$ \\
CFGen
& $0.625_{\pm.027}$
& $0.217_{\pm.011}$
& $0.0052_{\pm.0003}$
& $0.982_{\pm.001}$ \\
scANVI
& $0.625_{\pm.018}$
& $0.196_{\pm.009}$
& $0.0056_{\pm.0008}$
& $0.981_{\pm.004}$ \\
scVI
& $0.482_{\pm.040}$
& $0.268_{\pm.002}$
& $0.0190_{\pm.0004}$
& $0.919_{\pm.002}$ \\
\midrule
\multicolumn{5}{l}{
\textbf{Neuronal}\quad
\emph{rare}, $3{,}168$ training cells
} \\
Real cells
& $0.982$
& $0.031$
& $0.0019$
& $0.993$ \\
\textbf{CRNDiff}
& $\mathbf{0.971}_{\pm.012}$
& $\mathbf{0.051}_{\pm.003}$
& $\mathbf{0.0102}_{\pm.0018}$
& $\mathbf{0.956}_{\pm.005}$ \\
MDLM
& $0.784_{\pm.063}$
& $0.059_{\pm.011}$
& $0.0120_{\pm.0030}$
& $0.944_{\pm.019}$ \\
CFGen
& $0.734_{\pm.004}$
& $0.063_{\pm.003}$
& $0.0130_{\pm.0003}$
& $0.943_{\pm.006}$ \\
scANVI
& $0.269_{\pm.019}$
& $0.079_{\pm.003}$
& $0.0478_{\pm.0027}$
& $0.719_{\pm.021}$ \\
scVI
& $0.535_{\pm.073}$
& $0.074_{\pm.003}$
& $0.0394_{\pm.0062}$
& $0.781_{\pm.037}$ \\
\bottomrule
\end{tabular}
\renewcommand{\arraystretch}{1}
\end{table}


\section{Discussion}

CRNDiff specifies count-space diffusion through reaction channels and propensity functions. In the birth--death instance studied here, closed-form transition kernels, Markov bridges, and factorial-cumulant dynamics support reverse sampling, terminal-time selection, and inference-time conditioning of a frozen generator. The heart-atlas results suggest that separating generator learning from sampler-level conditioning is useful for rare target populations. Across three
\replaced[id=YQ]{target cell-types}{heart targets},
CRNDiff improves conditional fidelity, preserves marker-level differential-expression structure,
supports downstream classification, and reduces genealogical concentration relative to FK steering.

These conclusions have boundaries: terminal noising uses second-order excess covariance, higher-order dependence may require additional diagnostics, and finite-particle, density-ratio, or target-discriminator errors can shift rare-target purity--fidelity trade-offs. Lower library-size variability and median Fano factor indicate that steering selects target populations within the learned count distribution rather than repairing dispersion mismatch (Appendix~\ref{app:hvg-generator}). Future work should test coupled reactions and structured-state extensions while preserving tractable kernels, bridges, and samplers~\citep{danos2007rule,behr2016stochastic}.


\section*{Reproducibility Statement}

Code, configuration files, and the metric summaries used to construct all tables are available at
\url{https://anonymous.4open.science/r/CRNDiff-EDCA/}.
Appendices~\ref{app:toy}--\ref{app:hvg-generator} document data preprocessing, model training, sampler settings, and evaluation protocols.

\section*{Large Language Model Statement}

Large language models and associated AI tools were used for the following:

\begin{enumerate}
    \item \textbf{Writing.}
    LLM-based tools were used to assist with the retrieval and discovery of related
    work and to polish portions of the manuscript for clarity and readability.
    All cited references and AI-assisted text were checked, reviewed, and revised
    by the authors.

    \item \textbf{Code.}
    AI coding assistants were used to assist with portions of code implementation
    and debugging. All AI-assisted code was reviewed and tested by the authors.

    \item \textbf{Not used.}
    LLMs were not used to develop the theoretical or conceptual framework,
    formulate or prove mathematical claims, propose the method or hypotheses,
    design the experiments, generate reported data, or interpret the experimental
    results.
\end{enumerate}

The authors take full responsibility for the final content of this work, including
all text, claims, code, and other artifacts produced with the aid of generative AI.

\bibliographystyle{iclr2027_conference}
\bibliography{iclr2027_conference}

\clearpage
\appendix

\section{Derivations}

\subsection{Reverse process, FFBS, and endpoint tilting}
\label{app:reverse}

\topic{Reverse master equation}
\label{sec:reverse_KFE}

Let \(r_s(x)=q_{T-s}(x)\). Differentiating with respect to reverse time,
\begin{align}
    \partial_s r_s(x)
    &=
    \sum_y
    \left[
        w(x\to y)r_s(x)
        -
        w(y\to x)r_s(y)
    \right]
    \nonumber\\
    &=
    \sum_y
    \left[
        \widetilde w_s(y\to x)r_s(y)
        -
        \widetilde w_s(x\to y)r_s(x)
    \right],
\end{align}
where
\[
    \widetilde w_s(y\to x)
    =
    w(x\to y)\frac{r_s(x)}{r_s(y)}.
\]
This gives Eq.~\eqref{eq:reverse_rates}.

Writing \(\St^r:=\St_{\cdot,r}\), the reverse of CRN reaction \(r\) changes
\(\n\to\n-\St^r\), with propensity
\begin{equation}
    \widetilde j_r(\n,s)
    =
    j_r(\n-\St^r)
    \frac{
        q_{T-s}(\n-\St^r)
    }{
        q_{T-s}(\n)
    }.
    \label{eq:reverse_propensity_CRN}
\end{equation}
Hence
\begin{equation}
    \partial_s\widetilde q_s(\n)
    =
    \sum_r
    \left[
        \widetilde j_r(\n+\St^r,s)
        \widetilde q_s(\n+\St^r)
        -
        \widetilde j_r(\n,s)
        \widetilde q_s(\n)
    \right],
    \qquad
    \widetilde q_s=q_{T-s}.
    \label{eq:reverse_master_CRN}
\end{equation}
Reverse propensities generally need not be mass action even when the
forward propensities are.

\topic{Exact backward sampling}
Substituting Eqs.~\eqref{eq:posterior}--\eqref{eq:bridge},
\begin{align}
    \sum_{x_0}
    Q(x_0\mid x_t,t)B(x_s\mid x_0,x_t)
    &=
    \frac{
        \overrightarrow q(x_t\mid x_s)
    }{
        q_t(x_t)
    }
    \sum_{x_0}
    \overrightarrow q(x_s\mid x_0)\mcl D(x_0)
    \nonumber\\
    &=
    \overrightarrow q(x_t\mid x_s)
    \frac{q_s(x_s)}{q_t(x_t)}.
\end{align}
Consequently,
\begin{equation}
    q_T(x_{t_K})
    \prod_{j=1}^K
    \overleftarrow q(
        x_{t_{j-1}}\mid x_{t_j}
    )
    =
    \mcl D(x_{t_0})
    \prod_{j=1}^K
    \overrightarrow q(
        x_{t_j}\mid x_{t_{j-1}}
    ),
    \label{eq:path_reversal_identity}
\end{equation}
which proves exact FFBS on any grid when initialized from \(q_T\).

\topic{Endpoint tilting}
If the endpoint distribution \(P\) is replaced by
\[
    P^H(x_0)
    =
    \frac{H(x_0)P(x_0)}
         {\sum_{x'}H(x')P(x')},
\]
Bayes' rule immediately gives
\[
    Q^H(x_0\mid x_t,t)
    =
    \frac{
        H(x_0)Q(x_0\mid x_t,t)
    }{
        \sum_{x'}H(x')Q(x'\mid x_t,t)
    },
\]
proving Eq.~\eqref{eq:tilt_posterior}. The bridge is unchanged because
the forward transition process is unchanged.

\subsection{Generating functions and birth--death solution}
\label{app:pgf}

\topic{Mass-action PGF equation}
For \(f_\z(\n)=\z^\n\),
\(
\partial_t\mathbb E[f_\z(\n_t)]
=
\mathbb E[(Lf_\z)(\n_t)]
\)
gives
\begin{align}
    \partial_tg(\z,t)
    &=
    \sum_r
    \mathbb E
    \left[
        j_r(\n)
        \left(
            \z^{\n+\St^r}-\z^\n
        \right)
    \right]
    \nonumber\\
    &=
    \sum_r
    k_rV^{1-|\Sm_{\cdot,r}|}
    \left(
        \z^{\Sp_{\cdot,r}}-\z^{\Sm_{\cdot,r}}
    \right)
    \mathbb E
    \left[
        \prod_i\frac{n_i!}{(n_i-\Sm_{i,r})!}
        \z^{\n-\Sm_{\cdot,r}}
    \right].
\end{align}
Since
\(
\mathbb E\left[\prod_i\frac{n_i!}{(n_i-a_i)!} \z^{\n-\mbf{a}}\right]
=
\partial_\z^\mbf{a}g
\),
this proves Eq.~\eqref{eq:pgf-pde_CRN}.

\topic{Birth--death characteristics}
\phantomsection
\label{sec:solve_PGF_PDE}

For one species,
\begin{equation}
    \partial_tg
    =
    (z-1)(k^+-k^-\partial_z)g,
    \label{eq:single_species_pgf}
\end{equation}
with Lagrange--Charpit equations
\begin{equation}
    dt
    =
    \frac{dz}{k^-(z-1)}
    =
    \frac{dg}{k^+(z-1)g}.
    \label{eq:lagrange_charpit}
\end{equation}
Writing
\(
\beta_t=e^{-k^-t}
\)
and
\(
\mu=k^+/k^-
\),
the characteristic through \(z_0\) satisfies
\[
    z_0=1+\beta_t(z-1),
    \qquad
    \frac{dg}{g}=\mu\,dz,
\]
and therefore
\begin{equation}
    g_t(z)
    =
    e^{\mu(1-\beta_t)(z-1)}
    g_0\!\left(1+\beta_t(z-1)\right).
    \label{eq:single_species_pgf_solution}
\end{equation}
For independent birth--death coordinates this gives
\begin{equation}
    g_t(\z)
    =
    \exp\!\left[
        \sum_i
        \mu_i(1-\beta_{t,i})(z_i-1)
    \right]
    g_0\!\left(
        \bm1+
        \boldsymbol\beta_t\odot(\z-\bm1)
    \right),
    \label{eq:pgfsol}
\end{equation}
where $\odot$ is an element-wise or Hadamard product.
No factorization of \(g_0\) is required. Setting
\(g_0(\z)=\z^{\n_0}\) and extracting coefficients gives
Eq.~\eqref{eq:BD_kernel}.

Taking \(\Phi_t=\log g_t\) yields
\begin{equation}
    \Phi_t(\z)
    =
    \sum_i
    \mu_i(1-\beta_{t,i})(z_i-1)
    +
    \Phi_0\!\left(
        \bm1+\boldsymbol\beta_t\odot(\z-\bm1)
    \right),
    \label{eq:birth_death_fcgf}
\end{equation}
from which Eq.~\eqref{eq:birth_death_cumulant} and the general
order-\(m\) factorial-cumulant decay follow by differentiation.

\subsection{Spectral terminal-time criterion}
\label{app:signal}

Under our parameter choice,
\(
\mud_t=\mud_0
\)
and
\(
\Psi_t=e^{-2t}\Psi_0
\).
Hence
\[
    \Sigma_t(\vv)
    =
    e^{-2t}
    \frac{\vv^\top\Psi_0\vv}
         {\vv^\top\mud_0\vv}.
\]
The stationary directions of this generalized Rayleigh quotient satisfy
\[
    \Psi_0\w_k
    =
    \sigma_k\mud_0\w_k,
\]
and therefore
\(
\Sigma_t(\w_k)=e^{-2t}\sigma_k
\).
Equivalently, the \(\sigma_k\) are the eigenvalues of
\[
    C
    =
    \mud_0^{-1/2}
    \Psi_0
    \mud_0^{-1/2}.
\]
Thus the strongest positive second-order signal is
\(e^{-2t}\sigma_1\), and equating it to the empirical Poisson threshold
\(\sigma_{\mathrm{noise}}\) gives
Eq.~\eqref{eq:optimal_terminal_time}.
{Under the product-Poisson null the excess covariance vanishes in population, so
\(\sigma_{\mathrm{noise}}\) is the largest normalized eigenvalue that a sample of \(N\) cells over \(S\)
genes produces by fluctuation alone. We take it to be the Marchenko--Pastur edge \citep{marchenko1967distribution}
\begin{equation}
    \sigma_{\mathrm{noise}}
    =
    \bigl(1+\sqrt{S/N}\bigr)^{2}-1 ,
    \label{eq:mp_edge}
\end{equation}
a closed-form function of \(S\) and \(N\) that involves no sampling.}
This criterion calibrates the decay of detectable positive excess-covariance structure; it does not by itself provide a bound on the discrepancy between \(q_{T_O}\) and \(\pi\). For the independent birth--death model, factorial cumulants of order \(m\geq 2\) decay as \(e^{-mt}\), but the thresholding rule is tied to the strongest positive second-order mode.


\subsection{Conditional sampler details}
\label{app:conditional_details}

\topic{Marginal information limitation}
Suppose two target distributions \(P_1\) and \(P_2\) have identical
coordinate marginals. They then give identical marginal ratios \(u_i\)
in Eq.~\eqref{eq:marginal_ratio}, and hence identical tilted posteriors
in Eq.~\eqref{eq:factorized_weight} at every reverse step. Their
marginally tilted reverse processes are therefore identical, proving
Proposition~\ref{prop:wall}.

This does not imply that the resulting endpoint distribution is a product
distribution: the factors
\(Q_{\theta,i}(n_{0,i}\mid\n_t,t)\) share the full conditioning variable
\(\n_t\), and marginalizing over \(\n_t\) can induce dependence.

\topic{Residual density ratio}
For balanced binary classification between
\(\Dref\) and \(\Dtilt\), the population log-loss minimizer is
\[
    r^*(\n)
    =
    \frac{
        \Dref(\n)
    }{
        \Dref(\n)+\Dtilt(\n)
    }.
\]
Its odds therefore satisfy
\[
    \frac{r^*(\n)}{1-r^*(\n)}
    =
    \frac{\Dref(\n)}
         {\Dtilt(\n)},
\]
which gives Eq.~\eqref{eq:rho}.

\topic{Intermediate FK potentials}
For the marginally tilted reverse chain, an ideal continuation potential is
\begin{equation}
    \psi_j(\n_{t_j})
    =
    \mathbb E
    \left[
        \rho_\phi(\n_0)
        \mid
        \n_{t_j}
    \right],
    \qquad
    0<j<K.
    \label{eq:fk_continuation_model}
\end{equation}
We approximate it using \(J\) clean-state predictions,
\begin{equation}
    \widehat\psi_j(\n_{t_j})
    =
    \frac1J
    \sum_{\ell=1}^J
    \rho_\phi\!\left(
        \widetilde\n_0^{(\ell)}
    \right),
    \qquad
    \widetilde\n_0^{(\ell)}
    \sim
    \Qtilt(
        \cdot\mid\n_{t_j},t_j).
    \label{eq:twist-mc}
\end{equation}
With
\(
\widehat\psi_K\equiv1
\)
and
\(
\widehat\psi_0(\n_0)=\rho_\phi(\n_0)
\),
the incremental weights are
\begin{equation}
    G_j
    =
    \frac{
        \widehat\psi_{j-1}(\n_{t_{j-1}})
    }{
        \widehat\psi_j(\n_{t_j})
    }.
    \label{eq:fk_increment}
\end{equation}
Along a particle lineage the same realized potential is carried forward,
so the product telescopes:
\[
    \prod_{j=1}^K G_j
    =
    \rho_\phi(\n_0).
\]
Consequently the weighted endpoint distribution is
\[
    \Dtilt(\n)\rho_\phi(\n),
\]
up to normalization, giving Eq.~\eqref{eq:fk_endpoint_model}.
Intermediate approximations affect weight variance and resampling
efficiency; the terminal density-ratio correction determines the formal
endpoint target.

\topic{Sampling configuration}
For the single-cell experiments, the sampler uses \(K=32\) reverse steps, \(M=2n_{\mathrm{out}}\) particles to generate \(n_{\mathrm{out}}\) cells, and \(J=16\) endpoint draws per particle for the intermediate potentials, taken
from the tilted posterior of the same step through a separate random stream so that the proposal
trajectory is unchanged. Resampling is systematic and is triggered when \(\ESS/M<0.5\) (\(\vartheta=0.5\) in Algorithm~\ref{alg:tilted_fk}), subject to the two-step minimum gap listed in Table~\ref{tab:hparams}, with
\begin{equation}
    \ESS = \frac{\bigl(\sum_{m=1}^{M} w^{(m)}\bigr)^{2}}
                        {\sum_{m=1}^{M}\bigl(w^{(m)}\bigr)^{2}} .
\end{equation}
After all reverse steps, the reported \(n_{\mathrm{out}}\) cells are selected from the \(M\) terminal particles by weighted systematic resampling using the final weights. This terminal weighted output selection is separate from the chain-internal resampling operations triggered by the ESS rule. Diagnostic resampling-event counts exclude the terminal output selection unless explicitly stated otherwise. The tempering exponent and the remaining settings are listed in Table~\ref{tab:hparams}.

\subsection{Connection to diffusion models}
\label{sec:diffusion_connection}

The CRN construction contains ordinary continuous diffusion as its
large-system-size approximation. Using
\(
e^{\Delta\cdot\partial_n}f(n)\equiv f(n+\Delta)
\),
the chemical master equation can be written
\[
    \partial_tq_t(n)
    =
    \sum_r
    \left(
        e^{-\St^r\cdot\partial_n}-1
    \right)
    j_r(n)q_t(n).
\]
Set \(x=n/V\), and let \(p_t(x)\) denote its density in the continuous approximation. For mass action,
\(j_r(Vx)=Va_r(x)+O(1)\), with
\(a_r(x)=k_rx^{\Sm_{\cdot,r}}\). Expanding to second order in \(V^{-1}\)
gives
\begin{equation}
    \partial_tp_t
    =
    -\nabla\cdot(Fp_t)
    +
    \frac{1}{2V}
    \sum_{i,j}
    \partial_{x_i}\partial_{x_j}
    (D_{ij}p_t)
    +
    O(V^{-2}),
    \label{eq:2nd_order_fwd}
\end{equation}
where
\[
    F(x)=\sum_r\St^ra_r(x),
    \qquad
    D(x)=\sum_r
    \St^r(\St^r)^\top a_r(x).
\]
Equivalently,
\begin{equation}
    dX_t
    =
    F(X_t)\,dt
    +
    V^{-1/2}\sigma(X_t)\,dW_t,
    \qquad
    \sigma=\sum_r \St^r\sqrt{a_r(x)}.
    \label{eq:forward_CLE}
\end{equation}

Write \(\rho_s(n)=q_{T-s}(n)\) for the reverse-time mass function, and use \(\rho_s(x)=p_{T-s}(x)\) for its continuous density approximation. Expanding the reverse propensities
\[
    \widetilde j_r(n,s)
    =
    j_r(n-\St^r)
    \frac{\rho_s(n-\St^r)}
         {\rho_s(n)}
\]
in the same way gives the reverse chemical Langevin equation
\begin{equation}
    dX_s
    =
    \left[
        -F(X_s)
        +
        \frac1V
        \left(
            \nabla\cdot D(X_s)
            +
            D(X_s)\nabla\log\rho_s(X_s)
        \right)
    \right]ds
    +
    V^{-1/2}\sigma(X_s)\,dW_s,
    \label{eq:reverse_CLE}
\end{equation}
where
\[
    \nabla f(X_s)
    :=
    \left.\nabla_x f(x)\right|_{x=X_s},
    \qquad
    (\nabla\!\cdot D)_i
    :=
    \sum_{j=1}^S \partial_{x_j}D_{ij}.
\]
Thus the large-\(V\) reverse CRN has the familiar reverse-diffusion
structure. At finite \(V\), the corresponding reaction-resolved
log-density difference is
\begin{equation}
    f_r(n,s)
    =
    \log
    \frac{
        \rho_s(n-\St^r)
    }{
        \rho_s(n)
    },
    \label{eq:discrete_reaction_score}
\end{equation}
so
\(
\widetilde j_r(n,s)
=
j_r(n-\St^r)e^{f_r(n,s)}
\).
Our sampler uses exact count kernels at \(V=1\), so neither the forward nor the reverse diffusion approximation is required.

\renewcommand{\topic}[1]{\par\Needspace{6\baselineskip}\smallskip\noindent\textbf{#1.}\ }
\setlength{\intextsep}{10pt plus 2pt minus 2pt}
\Needspace{20\baselineskip}
\section{Additional experimental results}
\label{app:experiments}

\Needspace{10\baselineskip}
\subsection{Terminal-time calibration}
\label{app:horizon_illustration}

\begin{figure}[H]
\centering
\includegraphics[width=.42\linewidth]{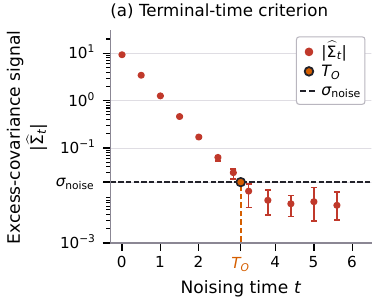}\hspace{.025\linewidth}%
\includegraphics[width=.42\linewidth]{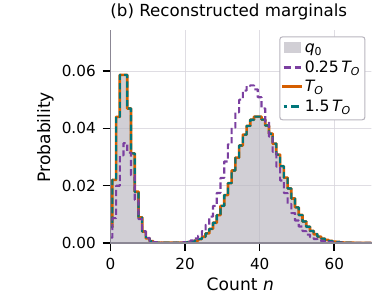}
\caption{Terminal-time illustration. (a) The estimated excess-covariance
signal reaches the empirical Poisson resolution floor at \(T_O=3.09\). (b)
reconstructed marginals obtained from three terminal times of the forward
process are compared with the data distribution in gray.}
\label{fig:t1d-criterion}\label{fig:t1d-marginals}
\end{figure}

\rev{Figure~\ref{fig:t1d-criterion} illustrates the terminal-time criterion of
Equation~\eqref{eq:optimal_terminal_time} on a one-dimensional example. The estimated
excess-covariance signal reaches the empirical Poisson resolution floor at
\(T_O=3.09\). The marginal reconstructed using terminal time \(0.25T_O\) differs visibly from the data distribution. Those obtained using \(T_O\) and \(1.5T_O\) are similar to the data and to each other. This example illustrates that the criterion selects a
terminal time beyond which additional forward noising produces little visible
change in the reconstructed marginal.}

\Needspace{10\baselineskip}
\subsection{Two-dimensional count mixture}
\label{app:toy}

\begin{figure}[H]
\centering\includegraphics[width=\linewidth]{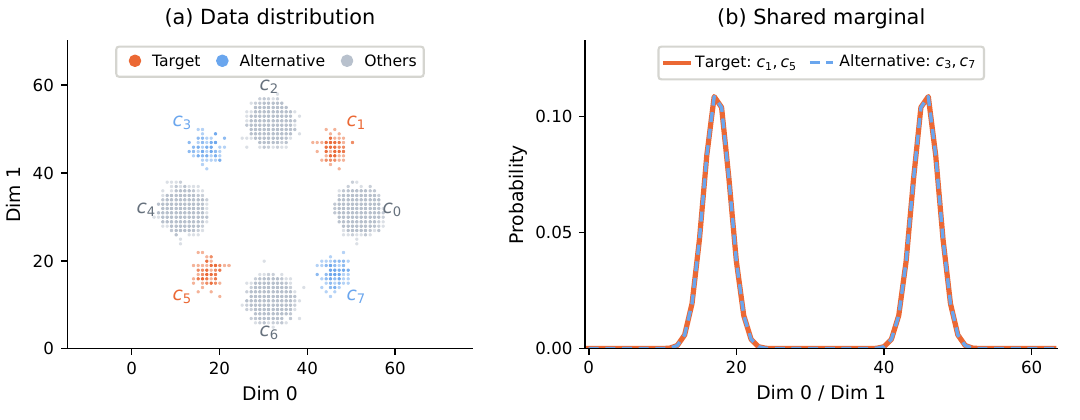}
\caption{Two-dimensional count mixture and matched one-dimensional marginals. Components \(\{c_1,c_5\}\) define the target distribution, and \(\{c_3,c_7\}\) define an alternative distribution with identical coordinate marginals. The component pairs occupy different regions of the count space.}
\label{fig:toy-construction}
\end{figure}

\paragraph{Construction.}
\label{app:toy-construction}
The eight-component mixture of Figure~\ref{fig:toy-construction} is defined on
the count lattice \(\mathcal G=\{0,\ldots,63\}^2\). Each component is a
discrete Gaussian kernel with
\[
p_j(\mathbf n)\propto
\exp\!\left(
-\frac{\|\mathbf n-\boldsymbol{\mu}_j\|_2^2}{2\sigma^2}
\right),
\qquad
\mathbf n\in\mathcal G,
\]
normalized over \(\mathcal G\), with \(\sigma=1.8\) and
\[
\boldsymbol{\mu}_j
=
(31.5,31.5)
+
20\bigl(\cos(j\pi/4),\sin(j\pi/4)\bigr),
\qquad j=0,\ldots,7.
\]
The four axis components \(\{c_0,c_2,c_4,c_6\}\) each have mixture mass
\(0.245\), while the four diagonal components
\(\{c_1,c_3,c_5,c_7\}\) each have mixture mass \(0.005\).
The target distribution assigns equal mass to \(c_1\) and \(c_5\), which
together account for \(1\%\) of the training mixture mass. The pair
\(\{c_3,c_7\}\) has the same one-dimensional marginals as the target pair to
numerical precision but a different joint distribution. This construction
therefore tests both sampling from a rare target distribution and recovery of
joint structure that is not determined by the marginals alone.

\rev{We compare three sampling configurations using the same frozen generator:
marginal tilt, FK steering, and tilted FK steering. All configurations use
\(M=10^4\) particles, output sample size \(n_{\mathrm{out}}=10^3\), and
\(K=32\) reverse steps. The two FK-steering configurations use \(J=16\)
posterior draws per intermediate potential. FK steering and tilted FK steering
resample when \(\ESS/M<0.5\).}

\paragraph{Distributional and genealogical metrics.}
For a generated distribution \(P\) and the target distribution \(P_R\), let
\(X\) and \(X'\) be independent samples from \(P\), and let \(Y\) and \(Y'\)
be independent samples from \(P_R\), with all four variables mutually
independent. The energy distance is
\begin{equation}
    \mathcal E(P,P_R)
    =
    2\mathbb E\|X-Y\|
    -\mathbb E\|X-X'\|
    -\mathbb E\|Y-Y'\|.
    \label{eq:toy-ed}
\end{equation}
It measures discrepancy between joint distributions without requiring a kernel
bandwidth and is related to a distance-induced MMD
\citep{szekely2013energy,sejdinovic2013equivalence}. The reported ED compares
generated samples with an equally sized i.i.d.\ sample from the exact target
distribution. Mean marginal total variation (mTV) averages, across the two
coordinates, the total-variation distance between the generated and target
one-dimensional marginals. Because the target pair \(\{c_1,c_5\}\) and the
alternative pair \(\{c_3,c_7\}\) have identical coordinate marginals, mTV alone
cannot distinguish them. Anc.\ denotes genealogical diversity, defined as the
number of distinct initial ancestors represented among the
\(n_{\mathrm{out}}=10^3\) generated samples divided by \(n_{\mathrm{out}}\).
Higher Anc.\ therefore indicates that the final output sample descends from
a larger set of initial particles. The i.i.d.\ reference provides a finite-sample
calibration for ED and mTV.
 
\begin{figure}[H]
\centering\includegraphics[width=\linewidth]{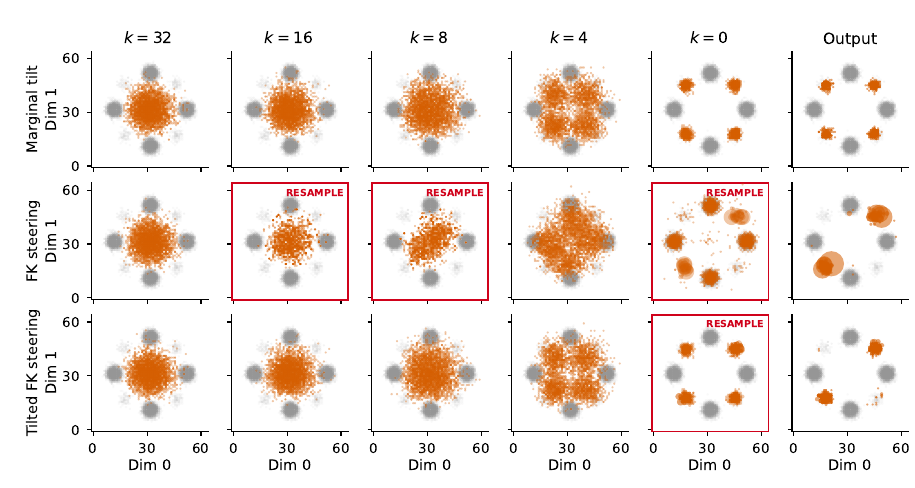}
\caption{Reverse-chain snapshots for marginal tilt, FK steering, and tilted FK
steering. Gray markers show training samples; orange markers show generated particles. In
the intermediate columns, orange-marker area scales with normalized particle
weight. In the output column, one marker per initial ancestor is placed at its last descendant in output-array order, with area proportional to that ancestor's number of output descendants. Red frames mark resampling
events. FK steering also resamples at \(k=24\), which is not shown, and a final
resampling is forced at \(k=0\) before output sampling.}
\label{fig:toy-reverse}
\end{figure}

\paragraph{Reverse-chain analysis.}
\label{app:toy-reverse}
\rev{Figure~\ref{fig:toy-reverse} compares the three sampling configurations at
matched reverse steps. Marginal tilt generates samples near all four rare
components, without distinguishing the target pair \(\{c_1,c_5\}\) from
\(\{c_3,c_7\}\), consistent with Proposition~\ref{prop:wall}. FK steering uses
an untilted proposal and progressively concentrates particles on the target
pair, but its final output contains many descendants of relatively few initial
particles. Tilted FK steering also concentrates on the target pair while
retaining more distinct initial ancestors in the output. The intermediate
snapshots and final genealogy therefore distinguish concentration on the target
pair from genealogical diversity.}
 
\begin{figure}[H]
\centering\includegraphics[width=.62\linewidth]{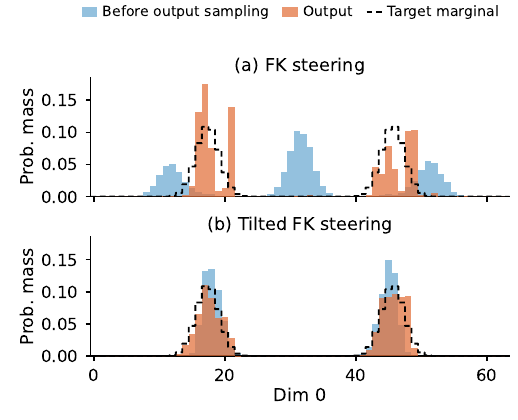}
\caption{Unweighted first-coordinate histograms before and after output sampling for FK steering and tilted FK steering, with the exact target marginal as a reference. Before output sampling, histograms use the endpoint particles; Output denotes the final samples drawn using the terminal weights. The comparison measures marginal agreement; the target and alternative component pairs have identical one-dimensional marginals.}
\label{fig:toy-resample}
\end{figure}

\rev{Figure~\ref{fig:toy-resample} compares unweighted first-coordinate histograms of the endpoint particles before output sampling with those of the output samples. The endpoint particles from tilted FK steering are already closer to the target marginal before output sampling, whereas FK steering requires a larger redistribution of probability mass.}

\Needspace{10\baselineskip}
\subsection{Single-cell experimental details}
\label{app:implementation}

\paragraph{Experimental setup.}
The scRNA-seq dataset from the adult human heart is split into training, validation, and test sets in an 8:1:1 ratio, shared across methods. Checkpoints are selected
by validation loss. Test cells serve as the real-data reference for distributional evaluation and are used to evaluate downstream classifiers. Generated-cell purity is scored from fixed CellTypist label histograms of each generated pool. Real-cell reference purity is computed over all held-out test cells in the target population after mapping CellTypist labels to the same fixed target label sets used for generated cells. All
models use the same \(2{,}000\) highly variable genes. CRNDiff and MDLM cap
model-input counts at \(512\), while scVI, scANVI, and CFGen use uncapped counts. Metric-specific preprocessing is shared across methods: \(W_1\) uses raw counts, differential-expression analysis uses CP10K-normalized counts, and MMD\({}^2\),
PCC, and downstream classification use log-CP10K values. CP10K normalizes each cell to total count \(10^4\) over the modeled genes; log-CP10K then applies \(\log(1+x)\). The endothelial, myeloid, and neuronal populations contain \(80{,}463\), \(18{,}422\), and \(3{,}168\) training cells, respectively. Generated and test samples are matched in size for distributional evaluation.

\paragraph{Generator.}
The posterior \(Q_\theta\) is parameterized by a transformer with self-attention across gene tokens and a per-gene categorical count head (Table~\ref{tab:hparams}). Each token encodes the noisy count, gene identity, and time. Training uses the cross-entropy objective of Section~\ref{sec:CT_diff_model}, with noisy samples drawn from the exact forward kernel in Equation~\eqref{eq:BD_kernel}. Three independent training seeds produce the models evaluated in Table~\ref{tab:hvg_single_type_main}. Conditioning comparisons and ablations use a single frozen model from this set as the reference CRNDiff generator.

\paragraph{Marginal-ratio estimation.}
For each frozen CRNDiff generator, per-gene histograms on \(0,\ldots,512\) are
computed from all training cells of the target population and a fixed pool of
\(20{,}000\) unconditional samples from that generator. The estimated log-marginal ratio is
\(\log(\widehat p_i+0.5/N_{\mathrm{target}})
-\log(\widehat q_i+0.5/20{,}000)\), clipped to \(\pm\log(10^3)\),
where \(\widehat p_i\) and \(\widehat q_i\) are normalized target and generator histograms, and \(N_{\mathrm{target}}\) counts target training cells. Zero-frequency target bins retain the smoothed correction.
These log-ratio estimates are reused across sampling runs. At each reverse step, we add \(\tau\) times the estimated log-marginal ratio to the posterior logits and normalize the resulting posterior probabilities. We use \(\tau=0.4\) for endothelial cells and \(0.2\) for myeloid and neuronal cells.

\paragraph{Residual discriminator.}
For each target population, \(r_\phi\) is a gradient-boosted tree classifier.
Positive examples are real target cells from the training set; negative examples
are cells generated by the marginally tilted proposal. Genes are ranked on
log-CP10K fitting samples by
\[
s_g =
\frac{\lvert \bar z_{P,g}-\bar z_{Q,g}\rvert}
{\sqrt{(v_{P,g}+v_{Q,g})/2+\epsilon}},
\]
Here \(P\) and \(Q\) denote target and proposal fitting samples; \(\bar z_{P,g},\bar z_{Q,g}\) and \(v_{P,g},v_{Q,g}\) are their means and empirical variances in log-CP10K space. The positive constant \(\epsilon\) stabilizes the denominator.
The top \(256\) genes are retained, except for the reference neuronal
discriminator, which uses \(192\) genes. Training uses early stopping
(Table~\ref{tab:hparams}). The odds \(r_\phi/(1-r_\phi)\) provide a
feature-based approximation to the residual density ratio in
Equation~\eqref{eq:rho}. For FK steering, the discriminator is fitted using
cells generated by the untilted proposal. The residual discriminator is
distinct from the external CellTypist evaluation classifier.
\begin{table}[H]
\centering\small
\setlength{\belowcaptionskip}{5pt}
\renewcommand{\arraystretch}{1.08}
\caption{Generator, sampling, residual-discriminator, and downstream-classifier settings used in the single-cell experiments.}
\label{tab:hparams}
\begin{tabular}{@{}
    >{\raggedright\arraybackslash}p{0.38\linewidth}
    >{\raggedright\arraybackslash}p{0.58\linewidth}
@{}}
\toprule
Setting & Value \\
\midrule

\multicolumn{2}{l}{\emph{Generator}} \\
Architecture & Transformer encoder; self-attention across genes \\
Input encoding & Scaled/log-count features, gene embeddings, time MLP \\
Layers / width / attention heads & 3 / 128 / 4 \\
Output head & Categorical over counts \(0,\ldots,512\) \\
Optimizer & AdamW \\
Learning rate / weight decay & \(2\times10^{-3}\) / \(10^{-4}\) \\
Learning-rate schedule & 500-step warm-up, then cosine decay \\
Batch size / training steps & 384 / \(20{,}000\) \\
EMA decay & 0.999 \\
Checkpoint selection & Lowest validation loss \\
Forward rates &
\(k_i^-=1\), \(\mu_i=k_i^+=\widehat{\mathbb E}_{\mathcal D}[n_i]\),
using training-set means \\
Terminal time &
\(T_O=5.20\), determined by Equation~\eqref{eq:optimal_terminal_time} \\

\midrule
\multicolumn{2}{l}{\emph{Sampling}} \\
Reverse steps \(K\) & 32 \\
Particles \(M\) & \(2n_{\mathrm{out}}\), where \(n_{\mathrm{out}}\) is the output sample size \\
Endpoint draws per potential \(J\) & 16 \\
Resampling trigger & \(\ESS/M<0.5\), minimum gap 2 steps \\
Terminal output & weighted draw from terminal particles \\
Tilt exponent \(\tau\) & 0.4 (endothelial), 0.2 (myeloid, neuronal) \\
Discriminator probability floor & \(10^{-8}\), no upper clip on the odds \\

\midrule
\multicolumn{2}{l}{\emph{Residual discriminator (gradient-boosted trees)}} \\
Input genes & 256 selected genes (192 for the reference neuronal discriminator) \\
Learning rate / boosting rounds & 0.06 / at most 180 \\
Leaves / min.\ samples per leaf / \(L_2\) & 31 / 30 / 1.0 \\
Early stopping & 15\% internal validation split, patience 20 \\

\midrule
\multicolumn{2}{l}{\emph{Downstream classifier (multinomial logistic regression)}} \\
Penalty / regularization & \(L_2\) / \(C=1\) \\
Solver / maximum iterations & lbfgs / \(2000\) \\
Class weights & balanced \\

\bottomrule
\end{tabular}
\end{table}

\paragraph{Evaluation metrics.}
The metrics in Section~\ref{sec:hvg} use the same preprocessing and scoring rules across methods. CellTypist labels are mapped to the three target cell types using fixed label sets. \(W_1\) is computed on raw counts at matched sample sizes. \(\mathrm{MMD}^2\) uses the biased estimator in log-CP10K space, with a Gaussian kernel whose bandwidth is the median pairwise distance among test target cells. PCC uses the corresponding \(2{,}000\)-gene population mean vectors. Unconditional sliced-\(W_1\) uses \(512\) random projections (Table~\ref{tab:hvg_uncond_fidelity}) and is distinct from per-gene \(W_1\).

\paragraph{Downstream classifier.}
Classifier hyperparameters for the downstream-utility evaluation are listed in
Table~\ref{tab:hparams}; the training and test protocol is detailed in
Appendix~\ref{app:hvg}.

\paragraph{Baselines.}
scVI, scANVI, and CFGen use the same data split and metric-specific preprocessing as CRNDiff. For conditional generation, scVI samples a standard-normal latent variable and decodes it with the target cell type fixed as a batch covariate. scANVI samples its latent variable from a prior conditioned on the target cell type, then decodes counts. CFGen fixes the target cell type during latent flow sampling and count decoding. For these three conditional baselines, library sizes are sampled with replacement from real test cells of the target population; Table~\ref{tab:hvg_uncond_fidelity} separately reports library-size sources for unconditional generation. Because the released MDLM implementation is designed for text data, we implement its published masked-diffusion objective using the same transformer backbone architecture as CRNDiff, representing counts as discrete tokens with an additional mask symbol. The conditional MDLM is initialized from an unconditional checkpoint and trained with a cell-type embedding added to the time embedding; sampling fixes this label throughout progressive unmasking.

\paragraph{Experimental repetitions.}
Table~\ref{tab:hvg_single_type_main} reports three independently trained models per method, with one generation run from each trained model. Means and sample standard deviations are computed across these three model-level realizations. Tables~\ref{tab:hvg_ablation}--\ref{tab:hvg_tau} and Figure~\ref{fig:hvg-ancestry} use the frozen reference CRNDiff generator and repeat stochastic sampling three times. Differential-expression and downstream-classification results use five evaluation resamplings of fixed generated-cell pools; these repetitions quantify evaluation variation rather than generator-training variation, and their purity columns repeat the means from Table~\ref{tab:hvg_single_type_main}. For downstream classification, each evaluation resampling redraws training subsets without replacement from the fixed generated-cell and real training-cell pools, refits the logistic-regression classifier on those subsets, and evaluates it on the same held-out test set.

\subsection{Biological fidelity and downstream utility}
\label{app:hvg}

\rev{We complement the conditional-fidelity metrics with analyses of
low-dimensional localization, cell-type-specific differential-expression
structure, and the utility of generated cells for downstream classification.}

\begin{figure}[!htbp]
\centering
\includegraphics[width=\linewidth]{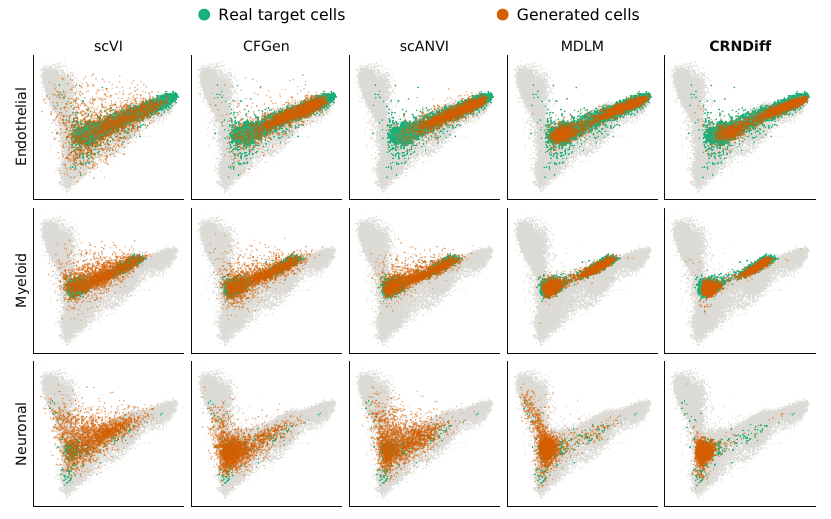}
\caption{Representative generated cells in the fixed atlas PCA basis, with one column per generative model and one row per target population in decreasing training abundance. Gray points show all test cells, with test cells from the target population highlighted in green; orange points show generated cells. Generated cells are subsampled to \(2{,}500\) cells per panel; the PCA basis is applied without refitting.}
\label{fig:hvg-pca-grid}
\end{figure}

\begin{figure}[!htbp]
\centering
\includegraphics[width=\linewidth]{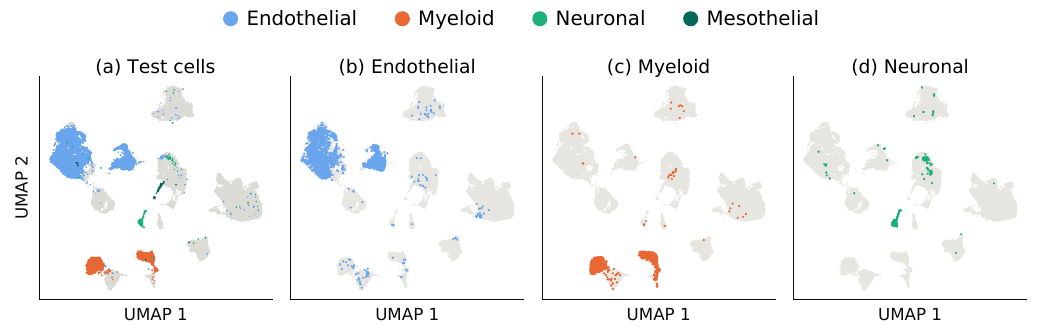}
\caption{CRNDiff-generated cells in the fixed atlas UMAP embedding. (a) Test cells (\(n=45{,}145\)); (b)--(d) \(2{,}500\) generated cells for endothelial, myeloid, and neuronal target populations. Mesothelial cells appear only in the test-set panel.}
\label{fig:hvg-umap}
\end{figure}

\paragraph{Fixed embeddings.}
\label{app:hvg-embeddings}
\rev{Figure~\ref{fig:hvg-pca-grid} compares representative generated cells from the five generative models in the fixed atlas PCA basis. Figure~\ref{fig:hvg-umap} shows CRNDiff-generated cells in the fixed atlas UMAP embedding. Both mappings place generated cells in the coordinate systems learned from real atlas cells, without refitting.}

\Needspace{8\baselineskip}
\rev{The PCA panels show that CRNDiff-generated neuronal cells concentrate near the real neuronal population, whereas generated cells from the baseline models show more dispersed projections. The CRNDiff UMAP panels provide a complementary view of target-population localization. These embeddings describe low-dimensional geometry and do not establish full count-space fidelity.}

\Needspace{8\baselineskip}
\paragraph{Differential-expression metrics.}
For each target cell type, we compare differential-expression rankings obtained from generated target cells and real test target cells. In both cases, the remaining cell types are represented by test cells. For each gene, log-fold change is the base-2 logarithm of the target-to-rest ratio of mean CP10K expression, with one added to each mean. Table~\ref{tab:hvg_de} reports overlap among the 100 genes with the largest signed log-fold changes and Spearman correlation across all \(2{,}000\) gene rankings. Results are averaged over five evaluation resamplings of fixed generated-cell pools; cells are not regenerated for these repetitions, and the real reference and test split remain fixed. These metrics assess differential-expression agreement rather than cell-type purity.

\begin{table}[H]
\centering\small
\setlength{\belowcaptionskip}{5pt}
\caption{Differential-expression agreement with test target cells. Center: overlap of the 100 genes with the largest signed target-versus-rest log-fold changes. Right: Spearman correlation across the rankings of all \(2{,}000\) genes. Subscripts report standard deviations over five evaluation resamplings of fixed generated-cell pools. Purity columns repeat the means from Table~\ref{tab:hvg_single_type_main} for comparison and are not recomputed across the five resamplings.}
\label{tab:hvg_de}
\setlength{\tabcolsep}{3pt}
\begin{adjustbox}{max width=\linewidth}
\begin{tabular}{lccccccccc}
\toprule
& \multicolumn{3}{c}{Purity (Table~\ref{tab:hvg_single_type_main})} & \multicolumn{3}{c}{Top-100 marker overlap $\uparrow$} & \multicolumn{3}{c}{Spearman, all 2,000 genes $\uparrow$} \\
\cmidrule(lr){2-4}\cmidrule(lr){5-7}\cmidrule(lr){8-10}
Model & Endo. & Mye. & Neu. & Endothelial & Myeloid & Neuronal & Endothelial & Myeloid & Neuronal \\
\midrule
scVI & \chg{$0.423$} & \chg{$0.482$} & \chg{$0.535$} & $0.806_{\pm.018}$ & $0.736_{\pm.015}$ & $0.692_{\pm.016}$ & $0.8923_{\pm.0032}$ & $0.9228_{\pm.0036}$ & $0.7234_{\pm.0110}$ \\
CFGen & \chg{$0.797$} & \chg{$0.625$} & \chg{$0.734$} & $0.932_{\pm.008}$ & $0.830_{\pm.012}$ & $0.812_{\pm.011}$ & $0.9812_{\pm.0011}$ & $0.9668_{\pm.0016}$ & $0.8735_{\pm.0018}$ \\
scANVI & \chg{$0.857$} & \chg{$0.625$} & \chg{$0.269$} & $0.932_{\pm.011}$ & $0.860_{\pm.021}$ & $0.742_{\pm.015}$ & $0.9772_{\pm.0004}$ & $0.9680_{\pm.0024}$ & $0.6930_{\pm.0091}$ \\
MDLM & \chg{$0.898$} & \chg{$0.840$} & \chg{$0.784$} & $0.902_{\pm.011}$ & $0.872_{\pm.019}$ & $0.842_{\pm.016}$ & $0.9767_{\pm.0026}$ & $0.9551_{\pm.0009}$ & $0.8003_{\pm.0027}$ \\
\textbf{CRNDiff} & \chg{$0.906$} & \chg{$0.886$} & \chg{$0.971$} & $0.944_{\pm.011}$ & $0.874_{\pm.018}$ & $0.820_{\pm.007}$ & $0.9852_{\pm.0007}$ & $0.9688_{\pm.0015}$ & $0.8355_{\pm.0020}$ \\
\bottomrule
\end{tabular}
\end{adjustbox}
\end{table}

\paragraph{Training on generated cells.}
An eleven-class multinomial logistic-regression cell-type classifier, with hyperparameters in Table~\ref{tab:hparams}, is trained on \(2{,}500\) cells per class in log-CP10K space and evaluated on all \(45{,}145\) held-out test cells. Generated cells replace real training cells for the three target classes; the remaining eight classes use real training cells. For each of the five evaluation resamplings, the training cells are redrawn without replacement from these fixed pools and the downstream classifier is refit; the generators are not retrained.

\begin{table}[H]
\centering\small
\setlength{\belowcaptionskip}{5pt}
\caption{Downstream classification with generated cells replacing real training cells for three target classes. \(F_1\) scores are means and standard deviations over five evaluation resamplings; macro \(F_1\) is the unweighted mean over eleven atlas classes. The real-cell reference trains all classes on real cells. Purity columns reproduce the external CellTypist means from Table~\ref{tab:hvg_single_type_main} for context. All methods use the same genes and log-CP10K preprocessing.}
\label{tab:hvg_tstr}
\setlength{\tabcolsep}{5pt}
\begin{adjustbox}{max width=\linewidth}
\begin{tabular}{lccccccc}
\toprule
& \multicolumn{3}{c}{Purity (Table~\ref{tab:hvg_single_type_main})} & & \multicolumn{3}{c}{$F_1$ on test cells} \\
\cmidrule(lr){2-4}\cmidrule(lr){6-8}
Training source & Endo. & Mye. & Neu. & Macro $F_1$ (11 classes) $\uparrow$ & Endothelial & Myeloid & Neuronal \\
\midrule
scVI & \chg{$0.423$} & \chg{$0.482$} & \chg{$0.535$} & $0.915 \pm 0.003$ & $0.947 \pm 0.006$ & $0.879 \pm 0.018$ & $0.800 \pm 0.018$ \\
CFGen & \chg{$0.797$} & \chg{$0.625$} & \chg{$0.734$} & $0.935 \pm 0.001$ & $0.982 \pm 0.002$ & $0.916 \pm 0.005$ & $0.901 \pm 0.005$ \\
scANVI & \chg{$0.857$} & \chg{$0.625$} & \chg{$0.269$} & $0.895 \pm 0.000$ & $0.970 \pm 0.001$ & $0.940 \pm 0.003$ & $0.457 \pm 0.012$ \\
MDLM & \chg{$0.898$} & \chg{$0.840$} & \chg{$0.784$} & $0.923 \pm 0.002$ & $0.982 \pm 0.002$ & $0.936 \pm 0.007$ & $0.711 \pm 0.019$ \\
\textbf{CRNDiff} & \chg{$0.906$} & \chg{$0.886$} & \chg{$0.971$} & $0.941 \pm 0.001$ & $0.980 \pm 0.002$ & $0.943 \pm 0.003$ & $0.915 \pm 0.009$ \\
\midrule
\emph{Real cells} & \upd{\emph{0.944}} & \upd{\emph{0.929}} & \upd{\emph{0.982}} & $0.951 \pm 0.002$ & $0.987 \pm 0.001$ & $0.966 \pm 0.005$ & $0.960 \pm 0.005$ \\
\bottomrule
\end{tabular}
\end{adjustbox}
\end{table}

\rev{CRNDiff-generated training cells yield macro \(F_1\) and target-class \(F_1\) values close to the real-data reference. Differences among generative models are largest for neuronal \(F_1\), while endothelial \(F_1\) is similar for several methods. This comparison establishes utility for the evaluated classification task, rather than interchangeability between generated and real cells.}

\Needspace{10\baselineskip}
\subsection{Sampler diagnostics and ablations}
\label{app:hvg-sampler-diagnostics}

\rev{We next analyze the conditioning mechanism on a frozen CRNDiff generator,
focusing on the contributions of marginal tilt and FK correction,
genealogical diversity under resampling, and sensitivity to the
marginal-tilt exponent.}

\begin{figure}[!htbp]
\centering\includegraphics[width=\linewidth]{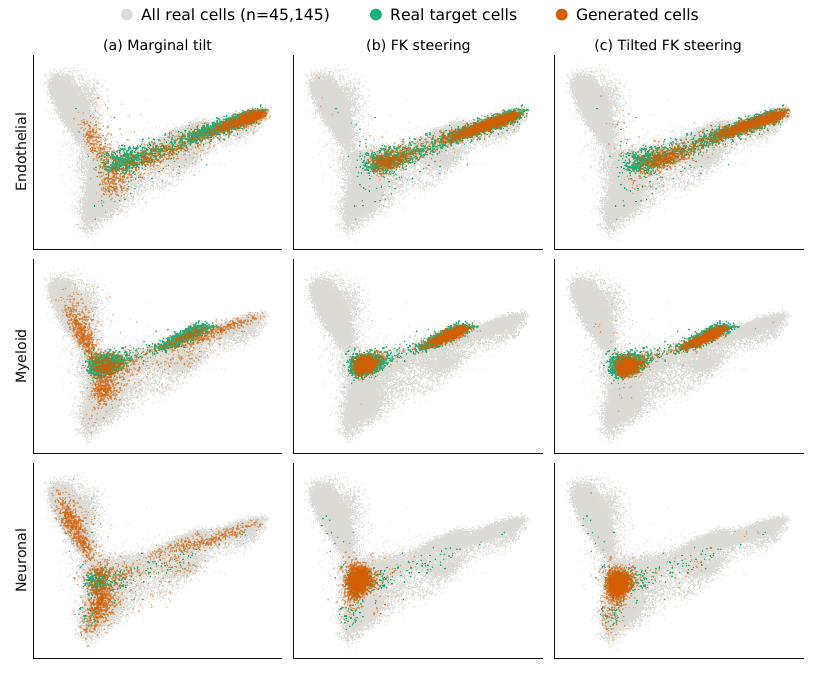}
\caption{Three sampling methods across the three target populations in the fixed atlas PCA basis: (a) marginal tilt, (b) FK steering, and (c) tilted FK steering. Colors follow Figure~\ref{fig:hvg-pca-grid}. Each panel shows \(2{,}500\) generated cells subsampled from the combined outputs of three sampling runs.}
\label{fig:hvg-ablation-pca}
\end{figure}

\paragraph{Conditioning-component ablation.}
\label{app:hvg-ablation}
\rev{Table~\ref{tab:hvg_ablation} and Figure~\ref{fig:hvg-ablation-pca} compare the conditioning components of Section~\ref{sec:fk} on one frozen CRNDiff generator. Marginal tilt adapts the posterior at every reverse step without particle resampling. FK steering uses the untilted proposal and a discriminator refitted against samples from that proposal. Tilted FK steering combines marginal tilt with particle weighting and resampling.}

\begin{table}[H]
\centering\small
\setlength{\belowcaptionskip}{5pt}
\caption{Ablation of marginal tilt and residual correction on the frozen reference CRNDiff generator, using the metrics of Table~\ref{tab:hvg_single_type_main}. The FK-steering discriminator is fitted against samples from the untilted proposal. Entries are means and standard deviations over three sampling runs.}
\label{tab:hvg_ablation}
\setlength{\tabcolsep}{3.5pt}\renewcommand{\arraystretch}{1.04}
\begin{adjustbox}{max width=\linewidth}
\begin{tabular}{lcccc}
\toprule
Sampler
& Purity $\uparrow$
& $W_1$ $\downarrow$
& MMD$^2$ $\downarrow$
& PCC $\uparrow$ \\
\midrule
\multicolumn{5}{l}{\textbf{Endothelial} \; \emph{abundant}, $80{,}463$ training cells} \\
Marginal tilt         & \upd{$0.682 \pm 0.000$} & $0.061 \pm 0.002$ & $0.0055 \pm 0.0002$ & $0.994 \pm 0.000$ \\
FK steering & \upd{$\mathbf{0.909} \pm 0.007$} & $\mathbf{0.043} \pm 0.009$ & $\mathbf{0.0005} \pm 0.0003$ & $\mathbf{0.999} \pm 0.000$ \\
Tilted FK steering & \upd{$0.906 \pm 0.002$} & $0.058 \pm 0.002$ & $0.0010 \pm 0.0001$ & $0.998 \pm 0.000$ \\
\midrule
\multicolumn{5}{l}{\textbf{Myeloid} \; $18{,}422$ training cells} \\
Marginal tilt         & \upd{$0.137 \pm 0.004$} & $0.297 \pm 0.002$ & $0.0504 \pm 0.0011$ & $0.764 \pm 0.008$ \\
FK steering & \upd{$\mathbf{0.889} \pm 0.016$} & $\mathbf{0.115} \pm 0.038$ & $0.0064 \pm 0.0048$ & $0.987 \pm 0.007$ \\
Tilted FK steering & \upd{$0.879 \pm 0.019$} & $0.131 \pm 0.007$ & $\mathbf{0.0045} \pm 0.0029$ & $\mathbf{0.991} \pm 0.008$ \\
\midrule
\multicolumn{5}{l}{\textbf{Neuronal} \; \emph{rare}, $3{,}168$ training cells} \\
Marginal tilt         & \upd{$0.063 \pm 0.005$} & $0.131 \pm 0.007$ & $0.0588 \pm 0.0003$ & $0.629 \pm 0.003$ \\
FK steering & \upd{$\mathbf{0.990} \pm 0.005$} & $0.062 \pm 0.009$ & $0.0179 \pm 0.0105$ & $0.923 \pm 0.043$ \\
Tilted FK steering & \upd{$0.984 \pm 0.002$} & $\mathbf{0.053} \pm 0.001$ & $\mathbf{0.0118} \pm 0.0003$ & $\mathbf{0.954} \pm 0.001$ \\
\bottomrule\end{tabular}
 
\end{adjustbox}
\end{table}

\rev{Marginal tilt yields dispersed PCA projections and low purity, particularly for the myeloid and neuronal target populations. These results indicate that adapting coordinate marginals is insufficient to recover the target population under the tested settings.}
 
\rev{Compared with tilted FK steering, FK steering gives better endothelial distributional metrics, while the myeloid results are mixed. Tilted FK steering improves the neuronal distributional metrics
(Table~\ref{tab:hvg_ablation}) and preserves greater genealogical diversity
for every evaluated target population (Table~\ref{tab:hvg_sampling_cost}), rather than uniformly improving every metric. Because the two procedures use separately fitted discriminators, their comparison also includes differences in density-ratio estimation.}

\paragraph{Genealogical diversity.}
\label{app:hvg-degeneracy}
\rev{When particle weights are concentrated, resampling favors descendants of a few particles and can reduce the number of distinct initial ancestors represented in the output. Stochastic transitions can subsequently produce distinct count vectors from descendants of the same ancestor, so output uniqueness and genealogical diversity measure different properties~\citep{jacob2015path}. Table~\ref{tab:hvg_sampling_cost} shows high output uniqueness alongside low genealogical diversity, particularly for neuronal FK steering. Distinct-ancestor counts diagnose shared particle histories but are not themselves estimates of effective sample size.}

\begin{table}[!htbp]
\centering\small
\setlength{\belowcaptionskip}{5pt}
\caption{Particle diagnostics on the frozen reference CRNDiff generator. Genealogical diversity (Anc.) is the number of distinct initial ancestors represented in the final output divided by \(n_{\mathrm{out}}\). Unique is the fraction of distinct output count vectors. \(\ESS_{\mathrm{pre}}/M\) is the median normalized ESS at intermediate resampling events. Entries are means and standard deviations over three sampling runs.}
\label{tab:hvg_sampling_cost}
\setlength{\tabcolsep}{4pt}\renewcommand{\arraystretch}{1.05}
\begin{adjustbox}{max width=\linewidth}
\begin{tabular}{lcccc}
\toprule
Sampler
& Resampling events
& Anc. $\uparrow$
& Unique $\uparrow$
& $\mathrm{ESS}_{\mathrm{pre}}/M$ $\uparrow$ \\
\midrule
\multicolumn{5}{l}{\textbf{Endothelial} \; \emph{abundant}, $80{,}463$ training cells} \\
FK steering & $11.3 \pm 0.6$ & $0.300 \pm 0.004$ & $0.950 \pm 0.023$ & $0.379 \pm 0.017$ \\
Tilted FK steering & $6.0 \pm 0.0$  & $0.533 \pm 0.005$ & $0.959 \pm 0.042$ & $0.469 \pm 0.010$ \\
\midrule
\multicolumn{5}{l}{\textbf{Myeloid} \; $18{,}422$ training cells} \\
FK steering & $11.3 \pm 0.6$ & $0.059 \pm 0.005$ & $0.920 \pm 0.014$ & $0.051 \pm 0.007$ \\
Tilted FK steering & $15.0 \pm 0.0$ & $0.112 \pm 0.010$ & $0.986 \pm 0.024$ & $0.147 \pm 0.004$ \\
\midrule
\multicolumn{5}{l}{\textbf{Neuronal} \; \emph{rare}, $3{,}168$ training cells} \\
FK steering & $14.3 \pm 1.5$ & $0.013 \pm 0.003$ & $0.936 \pm 0.042$ & $0.022 \pm 0.009$ \\
Tilted FK steering & $16.0 \pm 0.0$ & $0.045 \pm 0.004$ & $0.995 \pm 0.002$ & $0.053 \pm 0.002$ \\
\bottomrule
\end{tabular}
 
\end{adjustbox}
\end{table}

\begin{figure}[H]
\centering\includegraphics[width=\linewidth]{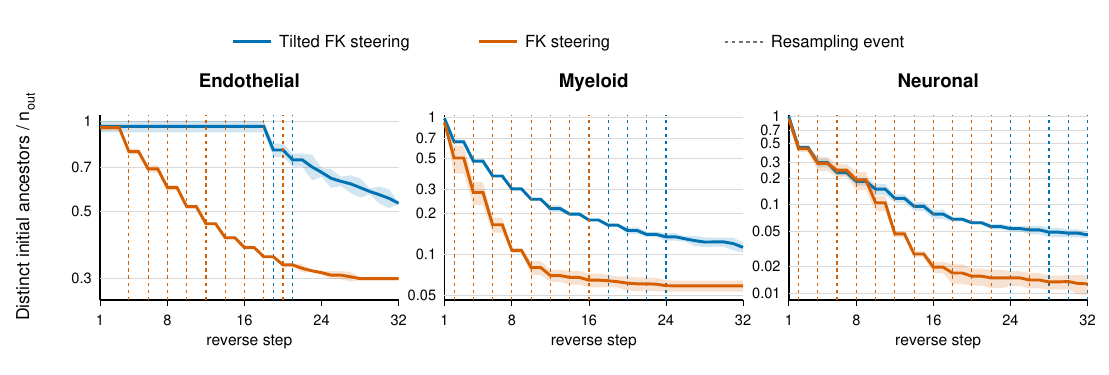}
\caption{Number of distinct initial ancestors represented along the reverse chain, normalized by output sample size. On logarithmic \(y\)-axes, curves show means and standard-deviation bands over three sampling runs. Dashed lines mark chain-internal resampling events shared across the three sampling runs, excluding the terminal weighted output selection. Final values correspond to Table~\ref{tab:hvg_sampling_cost}.}
\label{fig:hvg-ancestry}
\end{figure}

\rev{Despite more intermediate resampling events for myeloid and neuronal targets, tilted FK steering preserves greater final genealogical diversity (Table~\ref{tab:hvg_sampling_cost}; Figure~\ref{fig:hvg-ancestry}). Its higher median pre-resampling ESS is consistent with less concentrated weights at typical events. In these diagnostics, resampling-event counts and dashed markers refer to chain-internal resampling events and exclude the terminal weighted output selection described in Algorithm~\ref{alg:tilted_fk}. Pre-resampling ESS is measured after the incremental-weight update and may fall below the resampling threshold.}

\paragraph{Tempering the marginal tilt.}
\label{app:tilt-tempering}
\rev{The exponent \(\tau\) in Section~\ref{sec:fk} controls the strength of marginal proposal adaptation while leaving the target distribution unchanged. Table~\ref{tab:hvg_tau} evaluates the resulting purity and distributional fidelity on the frozen reference CRNDiff generator.}

\begin{table}[!htbp]
\centering\small
\setlength{\belowcaptionskip}{5pt}
\caption{Effect of the marginal tilt exponent, using the metrics of Table~\ref{tab:hvg_single_type_main}. The main experiments use \(\tau=0.4\) for endothelial cells and \(\tau=0.2\) for myeloid and neuronal cells. Entries are means and standard deviations over three sampling runs on the frozen reference CRNDiff generator.}
\label{tab:hvg_tau}
\setlength{\tabcolsep}{3.5pt}\renewcommand{\arraystretch}{1.04}
\begin{adjustbox}{max width=\linewidth}
\begin{tabular}{lcccc}
\toprule
$\tau$
& Purity $\uparrow$
& $W_1$ $\downarrow$
& MMD$^2$ $\downarrow$
& PCC $\uparrow$ \\
\midrule
\multicolumn{5}{l}{\textbf{Endothelial} \; \emph{abundant}, $80{,}463$ training cells} \\
$0.2$            & \upd{$0.906 \pm 0.005$} & $0.097 \pm 0.002$ & $0.0018 \pm 0.0001$ & $0.998 \pm 0.000$ \\
$0.4$ & \upd{$0.906 \pm 0.002$} & $0.058 \pm 0.002$ & $0.0010 \pm 0.0001$ & $0.998 \pm 0.000$ \\
$1.0$            & \upd{$0.933 \pm 0.004$} & $0.527 \pm 0.015$ & $0.1011 \pm 0.0027$ & $0.948 \pm 0.001$ \\
\midrule
\multicolumn{5}{l}{\textbf{Myeloid} \; $18{,}422$ training cells} \\
$0.2$ & \upd{$0.879 \pm 0.019$} & $0.131 \pm 0.007$ & $0.0045 \pm 0.0029$ & $0.991 \pm 0.008$ \\
$0.4$            & \upd{$0.796 \pm 0.006$} & $0.189 \pm 0.008$ & $0.0084 \pm 0.0010$ & $0.964 \pm 0.005$ \\
$1.0$            & \upd{$0.322 \pm 0.036$} & $0.839 \pm 0.065$ & $0.0923 \pm 0.0111$ & $0.914 \pm 0.003$ \\
\midrule
\multicolumn{5}{l}{\textbf{Neuronal} \; \emph{rare}, $3{,}168$ training cells} \\
$0.2$ & \upd{$0.984 \pm 0.002$} & $0.053 \pm 0.001$ & $0.0118 \pm 0.0003$ & $0.954 \pm 0.001$ \\
$0.4$            & \upd{$0.944 \pm 0.024$} & $0.059 \pm 0.000$ & $0.0106 \pm 0.0011$ & $0.958 \pm 0.006$ \\
$1.0$            & \upd{$0.912 \pm 0.045$} & $0.399 \pm 0.116$ & $0.0407 \pm 0.0092$ & $0.882 \pm 0.025$ \\
\bottomrule\end{tabular}
 
\end{adjustbox}
\end{table}

\Needspace{8\baselineskip}
\rev{The tested tempered settings improve all three distributional metrics relative to \(\tau=1\) across the three target populations. Purity and distributional fidelity do not always improve together: endothelial purity is highest at \(\tau=1\), despite worse distributional agreement. The trade-off between \(\tau=0.2\) and \(\tau=0.4\) depends on the target population and metric, so these results do not identify a single exponent that optimizes every outcome.}

\Needspace{10\baselineskip}
\subsection{Separating generator quality from conditioning}
\label{app:hvg-generator}\label{app:hvg-sampler}

\rev{We separate generator quality from conditioning by first evaluating
unconditional generation and then comparing conditioning strategies on a fixed
CRNDiff generator.}

\paragraph{Unconditional generation.}
For the \(2{,}000\) modeled genes, library size is the total count per cell, and detected-gene count is the number of nonzero entries. Their coefficients of variation (CVs) are sample standard deviations divided by means across cells. Median Fano is the median sample variance-to-mean ratio across genes with positive mean counts. These dispersion statistics use raw counts.

We evaluate unconditional fidelity and dispersion using the first \(20{,}000\) cells in each generated pool and an equally sized test sample (Table~\ref{tab:hvg_uncond_fidelity}). The CRNDiff generator has the
lowest sliced-\(W_1\) and the closest detected-gene coefficient of variation to
that of the test sample. Its median Fano factor and library-size coefficient of
variation are lower than the test-sample values. scANVI most closely matches the
median Fano factor, while CFGen most closely matches library-size variability.

CellTypist-assigned population fractions are computed over each full generated
pool and assess composition separately from distributional fidelity. The CRNDiff
generator most closely matches the test-sample endothelial and neuronal
fractions, while MDLM is closest for myeloid cells. Neuronal cells constitute
less than six percent of each generated pool; the number available for
conditional selection therefore depends on both this fraction and the candidate
budget.

\begin{table}[H]
\centering\small
\setlength{\belowcaptionskip}{5pt}
\caption{
Unconditional generation quality.
The library-size source indicates whether total library sizes are drawn from the training data or generated by the model.
Fidelity and dispersion metrics use the first \(20{,}000\) cells from each
generated pool. The final columns report cell-type fractions over the full
generated pool, assigned by the fixed external CellTypist heart model used for
purity, with the test sample shown as a reference. Entries are means over three
sampling runs per generator.}
\label{tab:hvg_uncond_fidelity}
\setlength{\tabcolsep}{3pt}
\begin{adjustbox}{max width=\linewidth}
\begin{tabular}{l l ccc c ccc}
\toprule
& Library & Median & Libsize & Det.\ genes & sliced-$W_1$
& \multicolumn{3}{c}{CellTypist-assigned cell-type fraction} \\
\cmidrule(lr){7-9}
Generator & size & Fano & CV & CV & $\downarrow$
& Endo. & Myeloid & Neuronal \\
\midrule
\emph{Test sample} & --- & 2.953 & 1.412 & 0.593 & ---
& \chg{\emph{0.233}} & \chg{\emph{0.0554}} & \chg{\emph{0.0211}} \\
\midrule
scVI       & drawn     & 2.625 & 1.403 & 0.718 & 0.0758 & \chg{0.194} & \chg{0.0403} & \chg{0.0555} \\
scANVI     & drawn     & 2.734 & 1.416 & 0.674 & 0.0345 & \chg{0.221} & \chg{0.0422} & \chg{0.0285} \\
CFGen      & drawn     & 3.922 & 1.411 & 0.698 & 0.0359 & \chg{0.212} & \chg{0.0438} & \chg{0.0401} \\
MDLM       & generated & 3.884 & 1.360 & 0.644 & 0.0723 & \chg{0.254} & \chg{0.0624} & \chg{0.0268} \\
\textbf{CRNDiff generator} & generated & 2.403 & 1.363 & 0.582 & \textbf{0.0234}
& \chg{0.229} & \chg{0.0412} & \chg{0.0247} \\
\bottomrule
\end{tabular}
 
\end{adjustbox}
\end{table}

\Needspace{6\baselineskip}
\paragraph{Conditioning a frozen generator.}
Table~\ref{tab:hvg_sampler_axis} compares best-of-\(N\) selection,
value-guided resampling~\citep{li2024derivativefreeguidancecontinuousdiscrete},
and tilted FK steering on one frozen CRNDiff generator. Best-of-\(N\) selects \(n_{\mathrm{out}}\) cells from \(N\) candidates ranked by target-class probabilities from a one-versus-rest gradient-boosted classifier trained on real training cells. A budget of \(b\times\) uses \(bB_1\) candidates, with \(B_1=39{,}599\) for endothelial cells and \(9{,}843\) for each other target, calibrated from generator forward-pass counts for the reference sampler.

\Needspace{5\baselineskip}
At each reverse step, value-guided resampling selects one of \(J=16\) candidates per particle with probability proportional to \(\widehat P_\phi(y\mid\mathbf n_s,s)^\gamma\), where \(y\) is the target class and \(\gamma\) controls selection strength. The time-conditioned classifier is trained by cross-entropy on forward-noised real training cells. A separate baseline fine-tunes the same generator on labeled cells.

\begin{table}[H]
\centering\footnotesize
\setlength{\belowcaptionskip}{5pt}
\caption{Conditioning strategies for the CRNDiff generator, using the metrics of Table~\ref{tab:hvg_single_type_main}. Entries are means over three sampling runs. The frozen-generator methods use the reference CRNDiff generator; \(\dagger\) denotes its conditionally fine-tuned counterpart. Candidate-pool sizes and guidance strengths do not imply equal computational cost.}
\label{tab:hvg_sampler_axis}
\setlength{\tabcolsep}{8pt}
\setlength{\arrayrulewidth}{0.6pt}
\renewcommand{\arraystretch}{0.95}
\newcommand{\grouprule}{\addlinespace[3pt]\cdashline{1-5}[3pt/2pt]\addlinespace[3pt]}
\begin{tabular}{l cccc}
\toprule
Conditioning & Purity $\uparrow$ & $W_1$ $\downarrow$ & MMD$^2$ $\downarrow$ & PCC $\uparrow$ \\
\midrule
\multicolumn{5}{l}{\textbf{Endothelial} \; \emph{abundant}, $80{,}463$ training cells; $n_{\mathrm{out}}=10{,}057$}\\
Best-of-$N$, $1\times$  & \upd{0.830} & 0.046 & 0.0012 & 0.998 \\
Best-of-$N$, $2\times$  & \upd{0.949} & 0.051 & 0.0033 & 0.998 \\
Best-of-$N$, $4\times$  & \upd{0.958} & 0.078 & 0.0079 & 0.996 \\
\grouprule
Value-guided, $\gamma=1$ & \upd{0.711} & 0.119 & 0.0117 & 0.992 \\
Value-guided, $\gamma=2$ & \upd{0.889} & 0.029 & 0.0006 & 0.999 \\
Value-guided, $\gamma=4$ & \upd{0.959} & 0.175 & 0.0169 & 0.994 \\
\grouprule
Tilted FK steering & \upd{0.906} & 0.058 & 0.0010 & 0.998 \\
\grouprule
Fine-tuned CRNDiff generator$^{\dagger}$ & \upd{0.929} & 0.043 & 0.0004 & 0.999 \\
\midrule
\multicolumn{5}{l}{\textbf{Myeloid} \; $18{,}422$ training cells; $n_{\mathrm{out}}=2{,}500$}\\
Best-of-$N$, $1\times$  & \upd{0.147} & 0.312 & 0.0577 & 0.690 \\
Best-of-$N$, $2\times$  & \upd{0.294} & 0.260 & 0.0343 & 0.878 \\
Best-of-$N$, $4\times$  & \upd{0.573} & 0.175 & 0.0104 & 0.976 \\
Best-of-$N$, $8\times$  & \upd{0.883} & 0.064 & 0.0012 & 0.996 \\
Best-of-$N$, $16\times$ & \upd{0.919} & 0.143 & 0.0093 & 0.988 \\
\grouprule
Value-guided, $\gamma=1$ & \upd{0.510} & 0.240 & 0.0156 & 0.964 \\
Value-guided, $\gamma=2$ & \upd{0.784} & 0.109 & 0.0019 & 0.992 \\
Value-guided, $\gamma=4$ & \upd{0.882} & 0.155 & 0.0150 & 0.973 \\
\grouprule
Tilted FK steering & \upd{0.879} & 0.131 & 0.0045 & 0.991 \\
\grouprule
Fine-tuned CRNDiff generator$^{\dagger}$ & \upd{0.897} & 0.106 & 0.0013 & 0.996 \\
\midrule
\multicolumn{5}{l}{\textbf{Neuronal} \; \emph{rare}, $3{,}168$ training cells; $n_{\mathrm{out}}=2{,}500$}\\
Best-of-$N$, $1\times$  & \upd{0.076} & 0.116 & 0.0612 & 0.610 \\
Best-of-$N$, $2\times$  & \upd{0.132} & 0.107 & 0.0542 & 0.664 \\
Best-of-$N$, $4\times$  & \upd{0.227} & 0.095 & 0.0457 & 0.731 \\
Best-of-$N$, $8\times$  & \upd{0.371} & 0.089 & 0.0366 & 0.795 \\
Best-of-$N$, $16\times$ & \upd{0.604} & 0.074 & 0.0230 & 0.883 \\
\grouprule
Value-guided, $\gamma=1$ & \upd{0.577} & 0.094 & 0.0250 & 0.882 \\
Value-guided, $\gamma=2$ & \upd{0.881} & 0.059 & 0.0077 & 0.965 \\
Value-guided, $\gamma=4$ & \upd{0.968} & 0.088 & 0.0116 & 0.956 \\
\grouprule
Tilted FK steering & \upd{0.984} & 0.053 & 0.0118 & 0.954 \\
\grouprule
Fine-tuned CRNDiff generator$^{\dagger}$ & \upd{0.961} & 0.032 & 0.0026 & 0.990 \\
\bottomrule
\end{tabular}
\end{table}

For endothelial cells, intermediate selection or guidance settings improve individual metrics relative to tilted FK steering, while stronger selection increases purity and distributional distances. For myeloid cells, best-of-\(N\) at \(8\times\) improves all three distributional metrics at similar purity; \(16\times\) increases purity but worsens fidelity.

\paragraph{Rare neuronal target population.}
Tilted FK steering achieves higher purity and lower \(W_1\) than every tested best-of-\(N\) and value-guided setting. Best-of-\(N\) retains a large purity gap even at the largest tested candidate-pool budget. Value-guided resampling narrows the gap, but increasing \(\gamma\) from \(2\) to \(4\) worsens all three distributional metrics. Some value-guided settings have lower MMD\({}^2\) and higher PCC than tilted FK steering. Fine-tuning further improves neuronal distributional fidelity, while tilted FK steering retains higher purity without retraining the generator.


 
\end{document}